\documentclass[10pt,twocolumn,letterpaper]{article}

\usepackage{iccv}              

\definecolor{iccvblue}{rgb}{0.21,0.49,0.74}
\usepackage[pagebackref,breaklinks,colorlinks,allcolors=iccvblue]{hyperref}
\usepackage[accsupp]{axessibility}
\usepackage{kotex}
\usepackage{graphicx}
\usepackage{multirow}
\usepackage{gensymb}
\usepackage{pifont}
\usepackage{amsmath}
\usepackage{algorithm}
\usepackage{algpseudocode}
\usepackage{multicol}
\usepackage{caption}
\usepackage{subcaption}
\usepackage{colortbl}
\usepackage{setspace}
\usepackage{amssymb}  
\usepackage{bbm}
\usepackage{afterpage}
\usepackage{colortbl}
\usepackage{booktabs}

\newcommand{\namesec}{Section}
\newcommand{\namefig}{Fig.}
\newcommand{\nametab}{Table}
\newcommand{\namealg}{Algorithm}
\newcommand{\ourmodels}{WAVE}

\title{WAVE: Warp-Based View Guidance for Consistent Novel View Synthesis \\ Using a Single Image}

\author{Jiwoo Park \quad Tae Eun Choi \quad Youngjun Jun \quad Seong Jae Hwang\thanks{Corresponding author}\\
\\
Yonsei University\\
{\tt\small \{wldn1677, teunchoi, youngjun, seongjae\}@yonsei.ac.kr}
}

\begin{document}

\twocolumn[{%
\renewcommand\twocolumn[1][]{#1}%
\maketitle
\vspace{-20pt}
\begin{center}
    \centering
    \captionsetup{type=figure}
    \includegraphics[width=\linewidth]{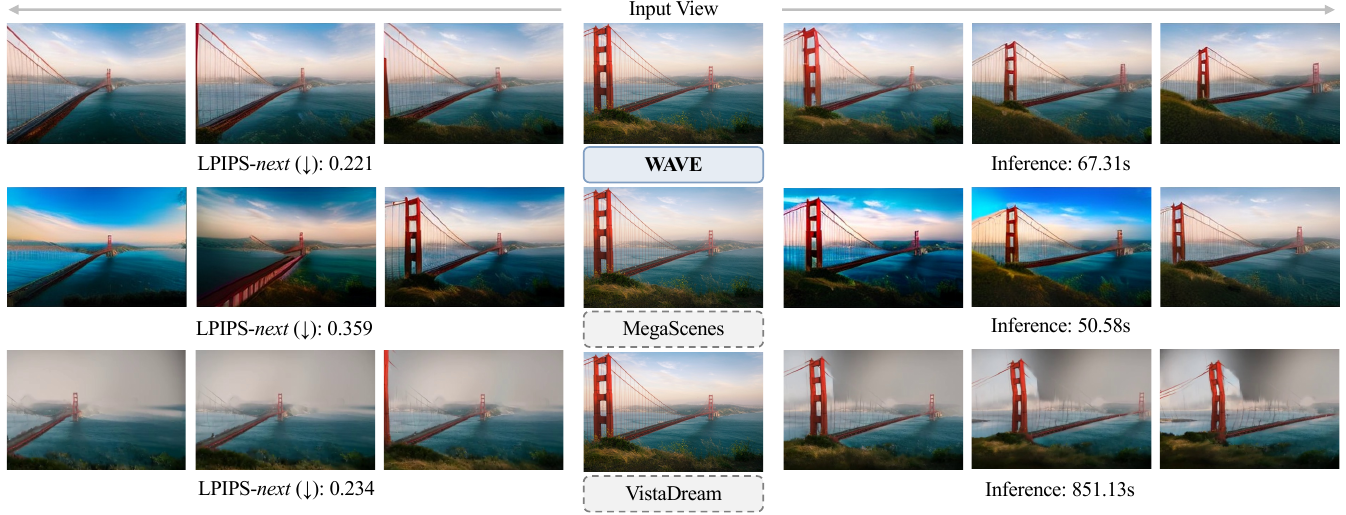}
    \vspace{-18pt}
    \captionof{figure}{\textbf{Teaser.} 
    We present WAVE, a training-free approach that enhances view consistency both among generated images and between the generated images and the input view. This approach is designed for novel view synthesis diffusion models using a input view. WAVE generates consistent images (lowest LPIPS-\textit{next}) across views without extra modules, enabling roughly 67s of inference time for 19 images.
    }\label{fig:teaser}
\end{center}%
}]
\begin{abstract}
\vskip -2mm
Generating high-quality novel views of a scene from a single image requires maintaining structural coherence across different views, referred to as view consistency.
While diffusion models have driven advancements in novel view synthesis, they still struggle to preserve spatial continuity across views. 
Diffusion models have been combined with 3D models to address the issue, but such approaches lack efficiency due to their complex multi-step pipelines.
This paper proposes a novel view-consistent image generation method which utilizes diffusion models without additional modules. 
Our key idea is to enhance diffusion models with a training-free method that enables adaptive attention manipulation and noise reinitialization by leveraging view-guided warping to ensure view consistency. 
Through our comprehensive metric framework suitable for novel-view datasets, we show that our method improves view consistency across various diffusion models, demonstrating its broader applicability
\footnote{\textit{Project page:} \href{https://jwoo-park0.github.io/wave.github.io/}{jwoo-park0.github.io/wave}}
\vskip -4mm

\end{abstract}

\vspace{-3pt}
\section{Introduction}
\vspace{-0.9pt}

\label{section:introduction}
Scene-level novel view synthesis from a single image has received significant interest for its potential to generate unseen views from only one image \cite{cat3d, wonderjourney, vista, infinitenature}. 
In single-image scenarios where multi-view information is unavailable, generative models have become a key component in novel view synthesis compared to rendering-based models \cite{3dgs,nerf}.
Specifically, diffusion models with strong generative capabilities have been adopted to generate novel views by utilizing view prior knowledge \cite{dreamgaussian, nerfdiff}.
Such approaches have led studies to explore diffusion models trained on large-scale datasets of images and camera parameters to generate novel views from a single image \cite{ZeroNVS, megascenes}. Leveraging their generative power, diffusion models can synthesize unseen scenes with zero-shot capability.

\begin{figure}[t!]
    \centering

    \includegraphics[width=\columnwidth]{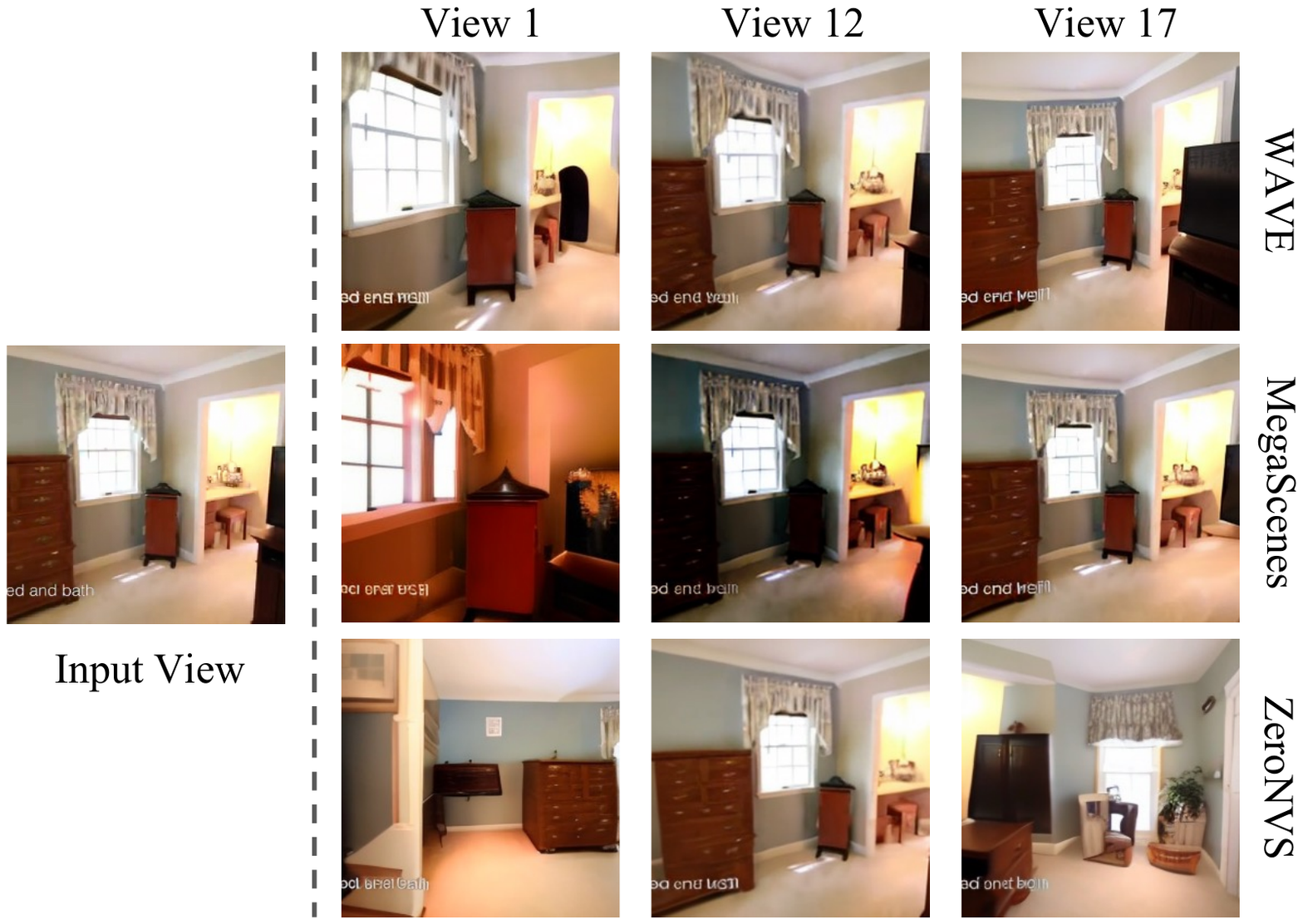} 
    \vspace{-15pt}

    \caption{\textbf{View inconsistency in diffusion models.} Generated novel views using the single image on the left are presented. WAVE maintains view consistency, whereas other diffusion-based models like MegaScenes \cite{megascenes} and ZeroNVS \cite{ZeroNVS} struggle to preserve consistency across viewpoints.}
    \vspace{-8pt}
    \label{figure_plus} 
\end{figure}

While diffusion-based models such as ZeroNVS \cite{ZeroNVS} and MegaScenes \cite{megascenes} have demonstrated strong performance, a key limitation of these models is that they cannot maintain view consistency among generated images from different viewpoints, as shown in \namefig{}~\ref{figure_plus}.
For example, MegaScenes shows view inconsistencies due to variations in object appearance and color saturation, as illustrated in the second row of \namefig{}~\ref{fig:teaser}.
To achieve view consistency, other studies \cite{vista,realmdreamer} have explored integrating diffusion models with additional models such as 3D models and large language models. 
In particular, VistaDream \cite{vista} leverages the 3D structures and contextual knowledge of vision language models to enhance consistency. 
Nevertheless, their applicability remains limited due to high computational costs and the complexity of multi-step pipelines. As shown in the third row of Fig.~\ref{fig:teaser}, despite the inference time being nearly ten times longer than that of MegaScenes, VistaDream still struggles to achieve satisfactory image quality.

Even with various efforts to complement diffusion models, inherent factors within diffusion models that compromise view consistency remain unresolved \cite{multidiff, genwarp}. 
A major factor is that existing diffusion models generate each image separately \cite{ZeroNVS, megascenes}, disregarding the relationship between images.
As a result, generated images lack interdependencies, making it challenging to maintain view consistency.
Additionally, diffusion models exhibit high sensitivity to noise randomness during the generation process \cite{prompt,freeinit,adapnoise}, leading to notable variations in the generated outputs. This variability impairs the model's ability to maintain overall visual coherence. 
Addressing these intrinsic factors directly could be a practical framework for achieving both view consistency and comparable inference time.
Therefore, we introduce a streamlined method, WAVE, for view consistency that addresses the challenges inherent in diffusion models without the need for extra modules or training. 

To strengthen image interdependencies, we propose \textbf{warp-guided adaptive attention}, which integrates 3D warping \cite{warp2} into batch self-attention to generate all viewpoints simultaneously. 
Batch self-attention computes attention by aggregating different keys and values across the batch, enabling information sharing between images \cite{storydiffusion, consisti2v,consistory}.
However, batch self-attention fails to account for viewpoint changes, limiting its ability to capture view-dependent information in novel view synthesis.
Thus, we utilize 3D warping which shifts pixels of an image from the input view to an arbitrary view. Integrated as a prior in batch self-attention, 3D warping provides cues for viewpoint transformations, improving view consistency.
Moreover, noise randomness introduces unintended variations in generated images, diminishing view consistency.
We propose \textbf{pose-aware noise initialization} which embeds the target image information into the initial diffusion noise to reduce noise randomness. Providing low-frequency information to the initial noise of diffusion models has proven effective in mitigating randomness \cite{freeinit, consisti2v}. 
Building on this, our method adopts the low-frequency of warped images to inject pose-aware information into the initial noise. As warped images inherently retain partial low-frequency details of the target view, they help alleviate noise randomness, ultimately enhancing view consistency.
 
In a evaluation of view consistency improvement, we construct the comprehensive view consistency metric framework. This framework combines video metrics \cite{storydiffusion,sparsectrl} and camera pose accuracy metrics as reconstruction-based metrics are unreliable for consistency evaluation \cite{vivid}.
Additionally, we evaluate image quality on a sequential dataset and validate our method in a downstream task using 3D Gaussian splatting \cite{3dgs}. The contributions are as follows:
\begin{itemize}
    \item We propose WAVE, a training-free method for scene-level novel view synthesis, adaptable to various diffusion models without additional training.
    
    \item We present two approaches to enhance view consistency: (1) warp-guided adaptive attention to facilitate inter-view information flow, and (2) pose-aware noise initialization to mitigate the negative impact of noise randomness. 
   
    \item Through our comprehensive metric framework along with extensive qualitative results, WAVE achieves improved performance over diffusion-based models, demonstrating comparable results to the rendering model.
   
\end{itemize}

\section{Related Work}
\label{section:related_works}

\subsection{Novel View Synthesis from Single Scene Image} 
Scene-level novel view synthesis from a single image has emerged as a crucial area of research, offering a feasible solution for generating realistic images on novel viewpoints while relying only on a single image. Unlike object-centric approaches \cite{Zero123, sv3d, Zero123++, objaverse} that focus on generating isolated objects, scene-level synthesis \cite{infinitenature,infinite, wonderjourney} aims to generate complex and complete environments.

Recent advancements in this domain have leveraged diffusion-based models, as they possess the generative capacity to generate novel views \cite{genwarp, multidiff,infinite}.
Consequently, efforts have been made to develop large-scale trained models \cite{megascenes, ZeroNVS} on Stable Diffusion \cite{ldm}. 
ZeroNVS \cite{ZeroNVS} introduces scene-scale variables for fine-tuning at scene level. MegaScenes \cite{megascenes} has since addressed the viewpoint inaccuracies and limited generalization ability of ZeroNVS by employing a warping strategy for scene images. 
Various works \cite{genwarp, multidiff} have also focused on enforcing consistency between the input view and the generated novel view images. 
Despite these contributions, ensuring robust consistency across multiple views synthesized by diffusion models remains a fundamental challenge.

To ensure view consistency, approaches such as VistaDream \cite{vista} and RealmDreamer \cite{realmdreamer} have integrated diffusion models with 3D models, reconstructing 3D from a single image or text inputs by leveraging 3D-aware representations. Although these approaches mitigate view inconsistency, they are made of multi-step pipelines that increases computational cost, making them less practical. 
While previous works have focused on auxiliary modules to assist diffusion models, the factors inherent to diffusion models which negatively impact view consistency remain unexplored.
Accordingly, a practical approach that resolves these inherent factors needs to be explored.

\subsection{Image Consistency in Diffusion Models} 
Diffusion models struggle to maintain consistency across image sequences in the fields of video generation \cite{consisti2v, storydiffusion}, text-to-image synthesis \cite{prompt, improving, consistory}, and image synthesis \cite{wu2023latent, condiffusion}.
Various methods have been proposed for consistency by leveraging attention or cross-view conditioning without additional training \cite{free3d, storydiffusion, consistory}. 
ConsiStory \cite{consistory} proposes a subject attention masking method to ensure consistent subject generation across different prompts, while StoryDiffusion \cite{storydiffusion} introduces an attention mechanism that enables the model to share contextual information. 

Other image-to-video generation methods \cite{freeinit, consisti2v, videobooth} have emphasized the negative effect of noise randomness on consistency, attributing this issue to the discrepancy between noise in the training and inference stages. FreeInit \cite{freeinit} introduces iterative sampling to mix low-frequency information from latent variables into the noise, while ConsistI2V \cite{consisti2v} incorporates low-frequency information from the first frame into the initial noise to mitigate randomness.
 
These methods have primarily focused on preserving structural continuity. In novel view synthesis, however, attempts to consistently maintain view-dependent spatial information have yet to be made.
Therefore, to ensure view consistency in novel view synthesis where view variation is a key factor, integrating geometric transformations for view changes into consistency methods is essential.

\section{Methods}
\label{section:methods}
\begin{figure}[t!]
    \centering
    \includegraphics[width=\columnwidth]{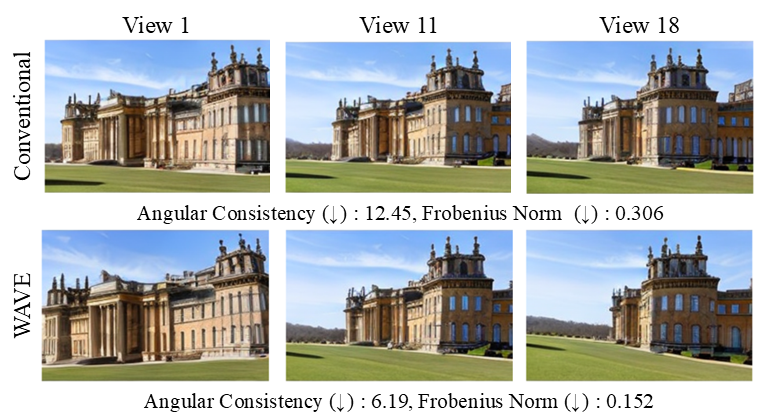}
    \vspace{-15pt}

    \caption{\textbf{Effectiveness of adaptive warp-range selection.} We present examples of applying our method alongside the conventional batch self-attention method \cite{free3d, consistory} that references all viewpoints. The differences in the two camera pose accuracy metrics, Angular Consistency and Frobenius Norm, show that the previous method generates images that are less accurately aligned with the camera poses than ours.}
    \label{figure_3} 
    \vspace{-8pt}
\end{figure}
 
\begin{figure*}[t!]
    \centering
    \includegraphics[width=\textwidth]{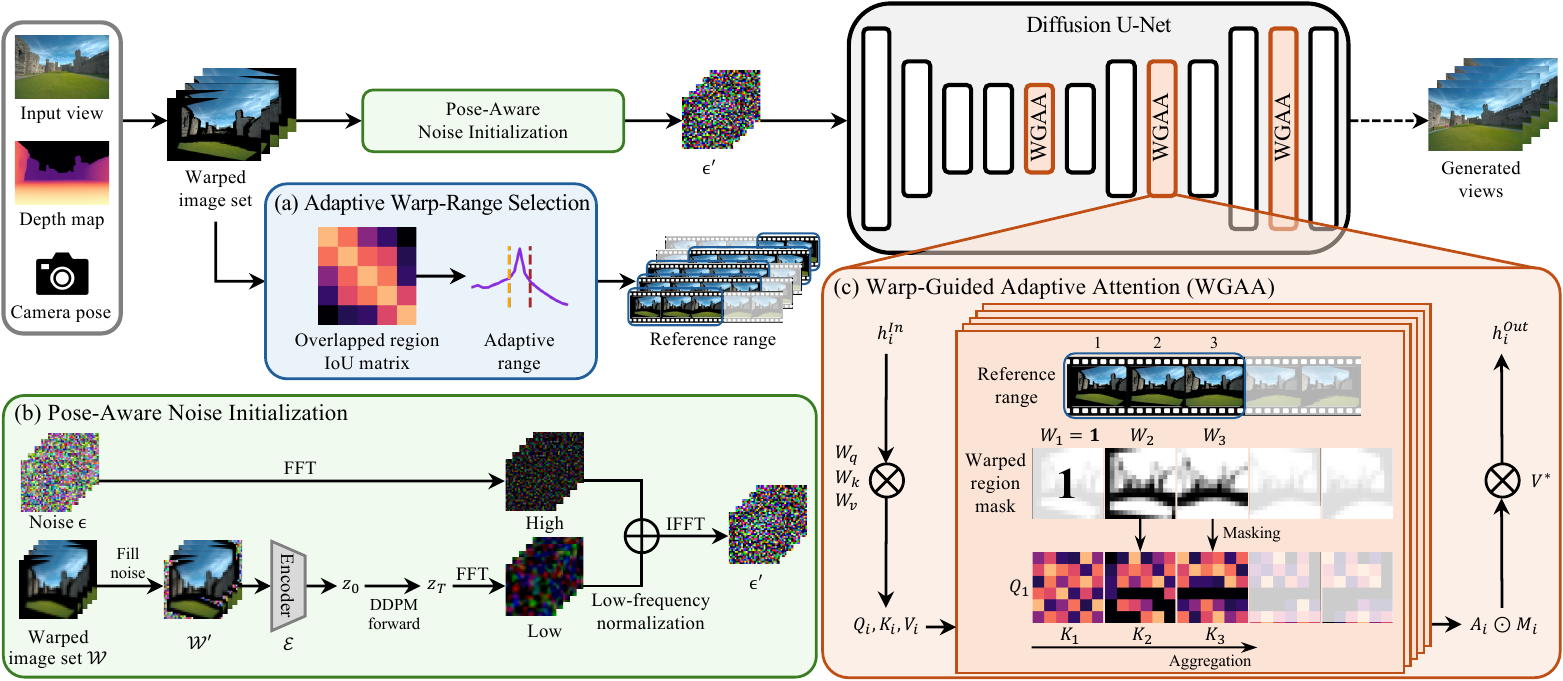} 
    \vspace{-15pt}
    \caption{\textbf{Illustration of WAVE.} Given an input view, depth map and continuous camera poses, our method generates scene images with smooth view transitions through three distinct processes: (a) adaptive warp-range selection utilizes warped region masks, from which the relevance between viewpoints is determined to compute the reference range for attention. (b) pose-aware noise initialization (PANI) re-initializes the diffusion initial noise by leveraging warped images and initial noise, incorporating frequency domain information. (c) warp-guided adaptive attention (WGAA) utilizes the warped region masks and reference range obtained from the adaptive warp-range selection, performing masked batch self-attention.}
    \label{figure_2} 
\vspace{-8pt}

\end{figure*}
We propose a training-free method that leverages warped images for consistent scene-level novel view synthesis using diffusion models pre-trained on large-scale datasets with camera poses and images \cite{ZeroNVS,megascenes}.
\subsection{Preliminaries}
\noindent\textbf{3D warping.}
3D warping \cite{warp2} is the process of transforming an input view image to match a desired camera pose using its depth. The transformation involves unprojecting image pixels into 3D space using depth values, followed by reprojecting them according to the specified camera poses. This approach enables the generation of warped images that accurately reflect the new viewpoints.

\vspace{3pt}
\noindent\textbf{Batch self-attention.}
Self-attention \cite{attention} extracts image features by applying linear projections with the attention matrices $W_q,W_k,W_v$ to produce $Q$, $K$, $V$. The attention map $A$ is then computed as $\mathtt{softmax}(QK^T / \sqrt{d_k})$ and multiplied by $V$ to generate hidden features $h$. Otherwise, batch self-attention \cite{free3d,consistory} aggregates the keys and values from all $N$ generations in a batch. Batch self-attention is defined as:
\small
\begin{align*} \label{eq:1}
     {Q_i} &\in \mathbb{R}^{p \times d_q}, \quad {K_i} \in \mathbb{R}^{p \times d_k}, \quad {V_i} \in \mathbb{R}^{p \times d_v}, \\
    {K^*} & = \big[{K_1, K_2, \ldots, K_N}\big] \in \mathbb{R}^{N \cdot p \times d_k}, \\
    {V^*} & = \big[{V_1, V_2, \ldots, V_N}\big] \in \mathbb{R}^{N \cdot p \times d_v}, 
    \\
    {A_i} & = \mathtt{softmax}({Q_i K^{*\top}}/\sqrt{d_k})\in \mathbb{R}^{p \times N \cdot p}, \\ {h_i}&= {A_i}\cdot {V^*} \in \mathbb{R}^{p \times d_v}. \tag{1}
\end{align*}
\normalsize
Batch self-attention enables the model to generate images from multiple viewpoints in a single forward pass, preventing diffusion models from generating images independently.
However, referencing all views with non-overlapping and unnecessary information reduces the reliance on the desired camera pose view features.
This makes it difficult to generate an image that matches the desired camera pose. 
To adequately reflect view variations, instead of referring to all views, we adaptively select the reference views (\namesec{}~\ref{sec:adaptive}) and develop an attention mechanism that masks non-overlapping regions within these selected views (\namesec{}~\ref{sec:warp_mask}).
\subsection{Adaptive Warp-range Selection}
\label{sec:adaptive}

Referencing all other viewpoints in the batch as in batch self-attention \cite{free3d,consistory} reduces viewpoint accuracy, as shown in Fig.~\ref{figure_3} and Supp. \nametab{}~\textcolor{iccvblue}{3}.
Fig.~\ref{figure_3} also shows examples of misalignment with the desired viewpoint.
Aggregating information from non-overlapping views ultimately degrades viewpoint accuracy.
Therefore, we set a reference range of viewpoints within the attention mechanism to utilize information from only a relevant subset of viewpoints. 

To determine the optimal reference range, we perform the adaptive warp-range selection shown in \namefig{}~\ref{figure_2}a. 
We utilize warped images from 3D warping to compute viewpoint relevance. 
As the viewpoint shifts, regions that are not visible from the input view become missing regions in the warped images, assigned a value of zero.
In order to create warped region masks $M$, we convert warped images into binary regions $W_{1:N}$, where pixels with a value of zero remain 0, and non-zero pixels are set to 1. 
Then, to measure the relevance between viewpoints (\eg, $i$ and $j$), we compute the IoU $u_{ij}$ on warped region masks to construct the matrix $U= \{u_{ij}\} \in \mathbb{R}^{N \times N}$, which provides a precise measure of their spatial relevance \cite{iou}. 
Based on this matrix, the adaptive range $[\mu_i-\sigma_i,\ \mu_i+\sigma_i]$ is selected for each viewpoint, for mean $\mu_i$ and standard deviation $\sigma_i$. Applying the standard deviation of the IoU matrix enables the dynamic determination of reference points, making our approach applicable across various sets of viewpoints.

\subsection{Warp-Guided Adaptive Attention}  \label{sec:warp_mask}
Moreover, in batch self-attention, as the number of referenced images increases, each image becomes overly dependent on information from other images rather than its own. To prevent excessive reliance on external information, previous work proposed a method of subject attention masking based on the subject mask to facilitate the selection of information \cite{consistory}. Attention masking allows the model to selectively focus on the important information. Building on this insight, we design warp-guided attention masking, which employs warped region masks in the diffusion decoder as shown in \namefig{}~\ref{figure_2}c, as follows: 
\vspace{-4pt}
\small
\begin{align*} \label{eq:mask}
    {A_i} &= \mathtt{softmax}({Q_iK^{*\top}/\sqrt{d_k}}),\\
    W_{1:N} &= \mathcal{T}(\mathcal{I}, \, \texttt{pose}_{1:N})\odot  \mathbf{1}_{\mathcal{T}(\mathcal{I}, \, \texttt{pose}_{1:N}) \neq 0}, \\
    M_i & = \big[W_1, W_2, \ldots, \textbf{1}, \ldots, W_N\big] , \\
    {h_i} & = ({A_i}\odot M_i) \cdot {V^*} \in \mathbb{R}^{p \times d_v},  \tag{2}
\end{align*}
\normalsize
where $\mathcal{T}$ is a 3D warping operation. This approach leverages the assumption that the decoder's attention mechanism encodes spatial information relevant to different views. For details of the assumption, please refer to the supplementary material. 
These warped region masks encapsulate the viewpoint cues of the target view image to be generated. The warped region masks effectively serve as a masking technique by capturing the spatial differences between viewpoints, allowing for a focused integration of important information.
Warp-guided adaptive attention performs warped region mask attention based on the adaptive range obtained through adaptive warp-range selection. This enables the attention mechanism to effectively reflect viewpoint changes.

\subsection{Pose-Aware Noise Initialization}  \label{sec:warp_init}
Beyond independent generation, noise randomness in diffusion models is another key issue. 
The stochastic nature of noise adversely impacts image consistency \cite{prompt,freeinit, adapnoise}, which extends similarly to view consistency.

View consistency, unlike image consistency, requires integrating information regarding view variations. 
Therefore, we propose pose-aware noise initialization which incorporates such information for noise reinitialization using warped images to enhance view consistency, as illustrated in \namefig{}~\ref{figure_2}b. 
Warped images capture the partial low-frequency of the target images, reflecting variations across different viewpoints. 
Leveraging them for noise reinitialization mitigates noise randomness by integrating geometric information from view changes into the noise.
The process is summarized in \namealg{}~\ref{algori_1}.
Warped images $\mathcal{W}$ are generated, aligned with the given camera poses. These images are passed through the VAE encoder $\mathcal{E}$, and DDPM noise is then added to make $z_T$. Low-frequency information is extracted from $z_T$ and mixed with the high-frequency components of the noise to preserve essential information in warped images while introducing randomness in the high-frequency parts to enhance visual details \cite{consisti2v, freeinit}. 

\begin{figure}[t!]
    \centering
    \includegraphics[width=\columnwidth]{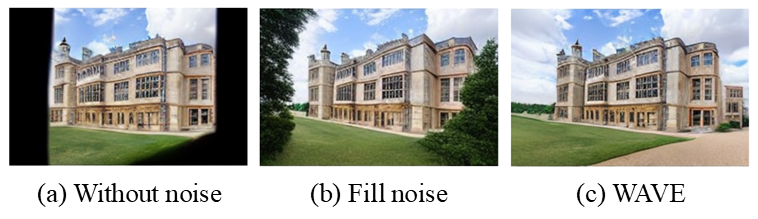} 
    \vspace{-20pt}
    \caption{\textbf{Analysis of warping methods.} We present the generation results of the methods attempted to mitigate the negative impact of missing regions in warped images : (a) not filling missing regions with noise, (b) only filling missing regions with noise, (c) our method (low-frequency normalization and filling noise).}
    \vspace{-8pt}
    \label{figure_4} 
\end{figure}
Warped images contain missing regions due to the absence of information from unseen viewpoints. 
Injecting low-frequency information directly from warped images results in undesirable black holes, as shown in Fig.~\ref{figure_4}a. 
This occurs because missing regions are reflected in the generated outputs. To address this, we apply random noise to fill the missing regions $\mathcal{W}'$ before injection, suppressing low-frequency associated with these regions. Otherwise, the low-frequency components of the random noise interfere with the original low-frequency of the warped images, leading to artifacts, as illustrated in Fig.~\ref{figure_4}b. To mitigate this, we normalize low-frequency components based on their magnitudes, controlling the impact of the random noise. This suppression of noise, as shown in Fig.~\ref{figure_4}c, ensures a more consistent generation process.

\begin{algorithm}[t!]
\caption{Pose-Aware Noise Initialization} \label{algori_1}
\begin{spacing}{1.25}
\begin{small}
\textbf{Input:} Reference image $\mathcal{I}$, camera poses $\texttt{pose}_{1:N}$, \\ 
\phantom{\textbf{Input:} } Depth map $\mathcal{D}$ \\
\textbf{Output:} re-initialized noise $\epsilon'$ \\
\textbf{Note:} Gaussian filter $\mathcal{G}(\mathcal{D}_0)$, VAE encoder $\mathcal{E}$ 
\begin{algorithmic}[1]
\State \text{Apply 3D Warping Algorithm using $\mathcal{I}$, $\mathcal{D}$, and $\texttt{pose}_{1:N}$} $\mathcal{W} \gets \{\mathcal{W}_1, \mathcal{W}_2, \dots, \mathcal{W}_n\}$
\State $\mathcal{W}' \gets \mathcal{W} + \eta \cdot \mathbf{1}_{W=0}$ \Comment{Add noise $\eta$ to zero regions of $W$}
\State $z_0 \gets \mathcal{E}(W)$ , \text{Sample} $\mathbf{\epsilon} \sim \mathcal{N}(0, I)$

\State $z_T \gets \text{DDPM Forward}(z_0,T)$  \Comment{noise scheduler}
\State $F_{T}^{\text{low}} \gets \text{FFT}(z_T) \odot \mathcal{G(D}_0)$
\State $F_{T}^{ \text{low}} \gets F_{T}^{\text{low}} / \text{mean}(|F_{T}^{\text{low}}|)$ \Comment{mean of $F_{T}^{\text{low}}$'s magnitude}
\State $F_{\epsilon}^{\text{high}} \gets \text{FFT}(\epsilon) \odot (1 - \mathcal{G(D}_0))$
\State $\epsilon' \gets \text{IFFT}(F_{T}^{ \text{low}} + F_{\epsilon}^{\text{high}})$
\State \textbf{Return} $\epsilon'$
\end{algorithmic}
\end{small}
\end{spacing}
\end{algorithm}
\section{Experiments}
\label{section:experiments}
\begin{table*}[t!]
\centering
\scriptsize
\setlength{\tabcolsep}{9pt} 
\renewcommand{\arraystretch}{1.1} 
\caption{\textbf{Consistency and camera accuracy metrics.} We conduct experiments on the three datasets MegaScenes, DTU, and RE10K and evaluate consistency using two metrics: \textit{Next} metric, which assesses consistency with neighboring viewpoints, and \textit{First} metric, which measures consistency with the input view. Additionally, to evaluate camera accuracy, a crucial aspect of novel view synthesis, we report the accuracy of camera extrinsic parameters using Frobenius Norm, Rotation Angle Difference, and Angular Consistency.}
\vspace{-5pt}

\resizebox{\textwidth}{!}{%
    \begin{tabular}{p{2.5cm} ccccccccc}
    
    \specialrule{0.7pt}{0pt}{0pt}
    
    \multirow{2}{*}{Method} & \multicolumn{2}{c}{\textit{Next}} & \multicolumn{2}{c}{\textit{First}} & {Frobenius Norm} & {Rotation Angle} & {Angular} \\
    \cline{2-3} \cline{4-5}
     & {LPIPS $\downarrow$} & {CLIPSIM $\uparrow$} & {LPIPS $\downarrow$} & {CLIPSIM $\uparrow$} & {(Rotation) $\downarrow$} & {Difference $\downarrow$} & {Consistency $\downarrow$} \\ 
     
    \specialrule{0.5pt}{0pt}{0pt}
    
    \multicolumn{8}{l}{\textbf{\textit{MegaScenes}}} \\ 
    ZeroNVS          & 0.561 & 0.887  & 0.602 & 0.826   & 0.563       & 0.410       & 23.53       \\
    \rowcolor[gray]{0.95} ZeroNVS + \ourmodels{}   & \textbf{0.404} & \textbf{0.933}  & \textbf{0.545} & \textbf{0.851} & \textbf{0.557}       & \textbf{0.404}       & \textbf{23.17}       \\ 
    MegaScenes & 0.607 & 0.861 & 0.609 & 0.813 & 0.481  & {0.355}  & {20.35} \\
    \rowcolor[gray]{0.95} MegaScenes + \ourmodels{}        & \textbf{0.397}  & \textbf{0.920}  & \textbf{0.561} & \textbf{0.826} & \textbf{0.382}   & \textbf{0.277}  & \textbf{15.91} \\
    
    \specialrule{0.4pt}{0pt}{0pt}
    
    \multicolumn{8}{l}{\textbf{\textit{DTU}}} \\ 
    ZeroNVS          & 0.489 & 0.883 & 0.726 & 0.717 & 0.568 & 0.414 & 23.77 \\
    \rowcolor[gray]{0.95} ZeroNVS + \ourmodels{}   & \textbf{0.344} & \textbf{0.925} & \textbf{0.704} & \textbf{0.739} & \textbf{0.550} & \textbf{0.401} & \textbf{23.00} \\ 
    MegaScenes & 0.379 & 0.898 & 0.692 & 0.697 & 0.308 & 0.227 & 13.04 \\
    \rowcolor[gray]{0.95} MegaScenes + \ourmodels{}        & \textbf{0.130} & \textbf{0.956} & \textbf{0.634} & \textbf{0.716} & \textbf{0.155} & \textbf{0.110} & \textbf{6.32} \\
    
    \specialrule{0.4pt}{0pt}{0pt}
    
    \multicolumn{8}{l}{\textbf{\textit{RE10K}}} \\ 
    ZeroNVS          & 0.478 & 0.897 & 0.764 & \textbf{0.745} & 0.624 & 0.459 & 26.30 \\
    \rowcolor[gray]{0.95} ZeroNVS + \ourmodels{}   & \textbf{0.388} & \textbf{0.921} & \textbf{0.761} & 0.743 & \textbf{0.620} & \textbf{0.455} & \textbf{26.10} \\ 
    MegaScenes & 0.347 & 0.912 & \textbf{0.722} & 0.676 & {0.242} & {0.179} & {10.25}\\
    \rowcolor[gray]{0.95} MegaScenes + \ourmodels{}        & \textbf{0.262} & \textbf{0.948} & 0.728 & \textbf{0.695} & \textbf{0.149} & \textbf{0.108} & \textbf{6.20} \\
    
    \specialrule{0.7pt}{0pt}{0pt}
    \end{tabular}
}

\vspace{-8pt}
\label{tab:main}
\end{table*}
\noindent\textbf{Baselines \& Datasets.} Our method is compared with other diffusion-based approaches, including MegaScenes \cite{megascenes} and ZeroNVS \cite{ZeroNVS}, as well as a multi-step pipelines model VistaDream \cite{vista}. Experiments conducted on the datasets MegaScenes \cite{megascenes}, RealEstate10K (RE10K) \cite{re10k}, DTU \cite{dtu}, and Mip-NeRF 360 \cite{mip} encompass synthetic, indoor, and outdoor scenes to ensure our method's generalizability. 

\vspace{3pt}
\noindent\textbf{Metric framework.} 
Several novel-view datasets \cite{megascenes, dtu} lack sequential viewpoints and ground-truth images, complicating the use of reconstruction metrics for consistency evaluation. The reliability of reconstruction metrics in evaluating consistency has been questioned \cite{vivid}. 
To address this, we design a view consistency metric framework that jointly evaluates image consistency and camera pose accuracy. This framework includes video metrics (LPIPS-\textit{next} / \textit{first} and CLIPSIM-\textit{next} / \textit{first}) alongside camera pose accuracy metrics (Frobenius Norm of rotation matrix, Rotation Angle Difference, and Angular Consistency).
For further details, please refer to Supp. Section \textcolor{iccvblue}{C}.

Experimental results based on our view consistency metric framework, along with results for reconstruction quality using the video dataset \cite{re10k} are presented in \namesec{}~\ref{main}. 
Additionally, \namesec{}~\ref{qualitative} presents qualitative comparisons against existing methods, \namesec{}~\ref{ablation} covers the ablation study, and \namesec{}~\ref{down} evaluates our method on a downstream task. Additional results and experimental settings can be found in the supplementary material.

\begin{figure}[t!]
    \centering
    \includegraphics[width=\columnwidth]{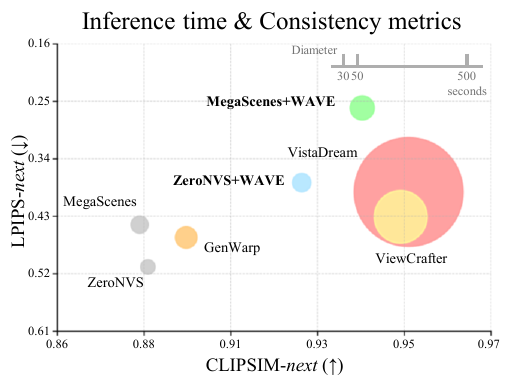} 
    \vspace{-18pt}
    \caption{\textbf{Overall comparison of methods.} The size of the circles represents inference time. Applying our method improves performance with comparable computation costs.}
    \vspace{-10pt}
    \label{figure_inference} 
\end{figure}

\subsection{Comparison with Baselines} \label{main}
\nametab{}~\ref{tab:main} compares novel view synthesis diffusion models with our method. 
Our method steadily outperforms diffusion-based models across diverse datasets, showing adaptability to diffusion models.
Specifically, on the MegaScenes dataset, our method achieves 0.397 in LPIPS-\textit{next} and 0.920 in CLIPSIM-\textit{next}, while generating images with higher accuracy in camera viewpoints. 
VistaDream\cite{vista} and Viewcrafter\cite{viewcrafter} incorporate a rendering method giving them an advantage in consistency metrics over diffusion models. Thus, these models are compared based on consistency metrics \textit{and} inference time, unlike other models. Additionally, we include Genwarp~\cite{genwarp} as a baseline to provide a more comprehensive comparison. Notably, as shown in \namefig{}~\ref{figure_inference}, the result of MegaScenes+WAVE exhibits significantly lower inference time than rendering-based methods while achieving comparable performance. 
This shows that solving the inherent factors of diffusion models can achieve a performance similar to rendering-based methods without excessive computational cost, demonstrating the practicality of our method.

Moreover, to evaluate image quality not captured by our metric framework, we employ the image sequence (\ie, video) dataset, RE10K, to assess the reconstruction and generation capabilities, as illustrated in \nametab{}~\ref{tab:re10k-seq-eval}. 
This dataset contains both ground-truth images and camera poses for varying views, allowing the evaluation of image quality.
The results show that our method outperforms existing models in both image generation (FID, KID) and reconstruction (PSNR, LPIPS) metrics.
The results demonstrate that our method produces consistent images while achieving higher image quality.

\begin{table}[t!]
\centering
\setlength{\tabcolsep}{9pt} 
\renewcommand{\arraystretch}{1.2} 
\caption{\textbf{RE10K sequence evaluation.} We present the results of image reconstruction and generation quality on RE10K, a video sequence dataset with varying viewpoints, demonstrating that our method outperforms the baseline.}
\vspace{-5pt}
\resizebox{\linewidth}{!}{
    \begin{tabular}{p{3.5cm} cccc}
    \specialrule{1pt}{0pt}{0pt}
    {Method} & {PSNR $\uparrow$} & {LPIPS $\downarrow$} & {FID $\downarrow$} & {KID $\downarrow$} \\
    \specialrule{0.7pt}{0pt}{0pt}
    VistaDream & 10.88 & 0.567 & 29.48 & 0.011 \\
    ZeroNVS          & 10.84   & 0.612    & {24.13}  & 0.010 \\
    \rowcolor[gray]{0.95} ZeroNVS + \ourmodels{}   & {11.38}  & {0.602}     & 25.67 & {0.009} \\
    MegaScenes & 12.02 & 0.469 & 18.20 & 0.006 \\
    \rowcolor[gray]{0.95} MegaScenes + \ourmodels{}        & \textbf{13.27} & \textbf{0.450}   & \textbf{15.78} & \textbf{0.004}  \\
    \specialrule{1pt}{0pt}{0pt}
    \end{tabular}
}

\label{tab:re10k-seq-eval}
\vspace{-8pt}
\end{table}

\begin{figure*}[t!]
    \centering
    \includegraphics[width=1\textwidth]{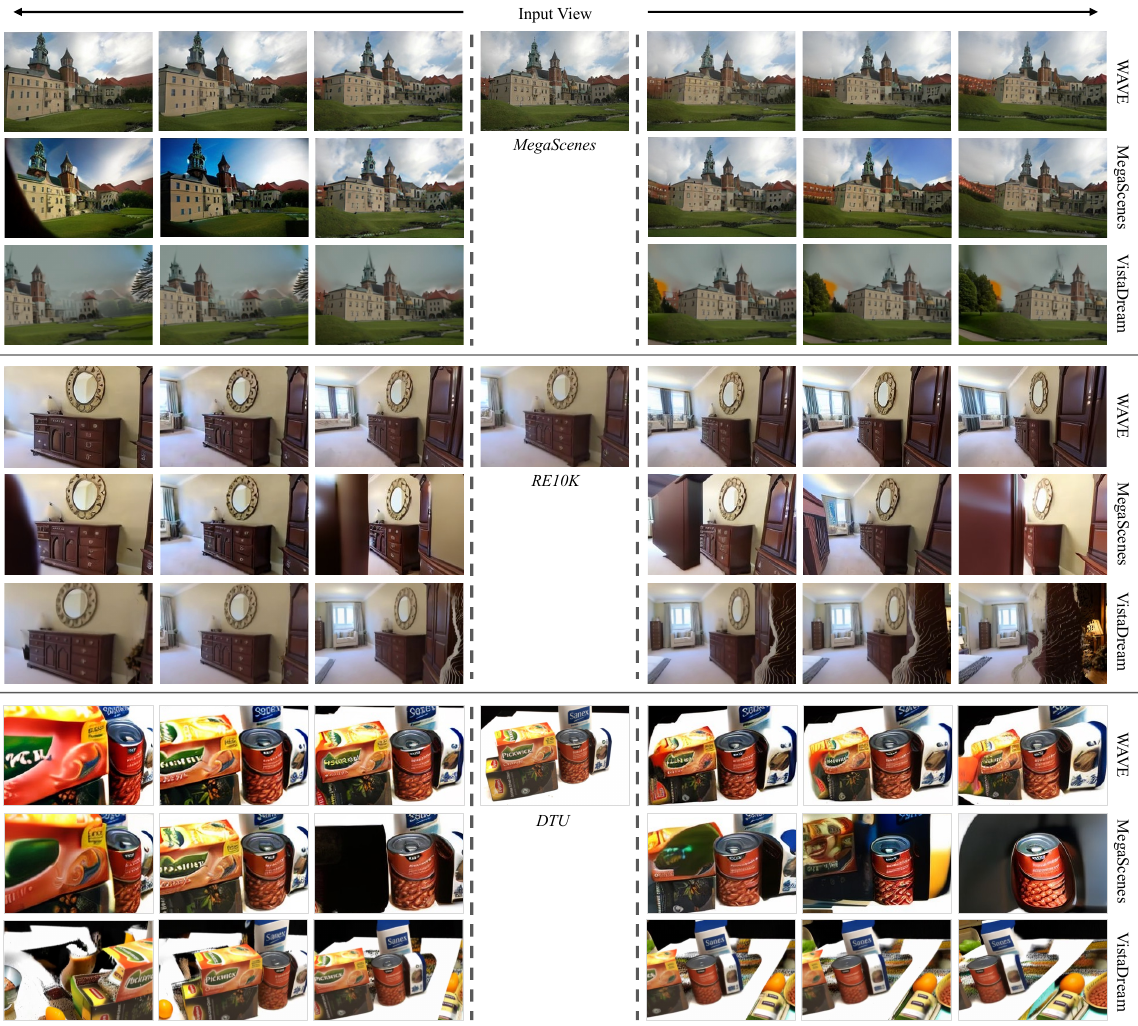}
    \vspace{-15pt}
    \caption{\textbf{Qualitative results.} We provide samples for our method and baseline models \cite{megascenes, ZeroNVS}. The results are obtained by generating images from an input view and continuous camera poses, selecting representative samples from them. Our method exhibits consistent images among generated images. The middle column presents the input view, while the remaining columns display the generated results.}
    \vspace{-8pt}
    \label{fig:qualitative} 
\end{figure*}

\subsection{Qualitative Comparison} \label{qualitative}
We compare our method against baselines with qualitative results. As shown in Fig.~\ref{fig:qualitative}, MegaScenes struggles to maintain consistency across different viewpoints. In contrast, WAVE produces more coherent results, addressing the limitations of existing diffusion models. 
While VistaDream maintains view consistency, it exhibits prominent artifacts in generated images. 
WAVE achieves superior performance compared to VistaDream, demonstrating that better-quality images can be generated without relying on multiple models.
More qualitative results and camera parameter visualization can be found in the supplementary materials.

\subsection{Ablation Study} \label{ablation}
\begin{figure}[t!]
    \centering
    \includegraphics[width=\columnwidth]{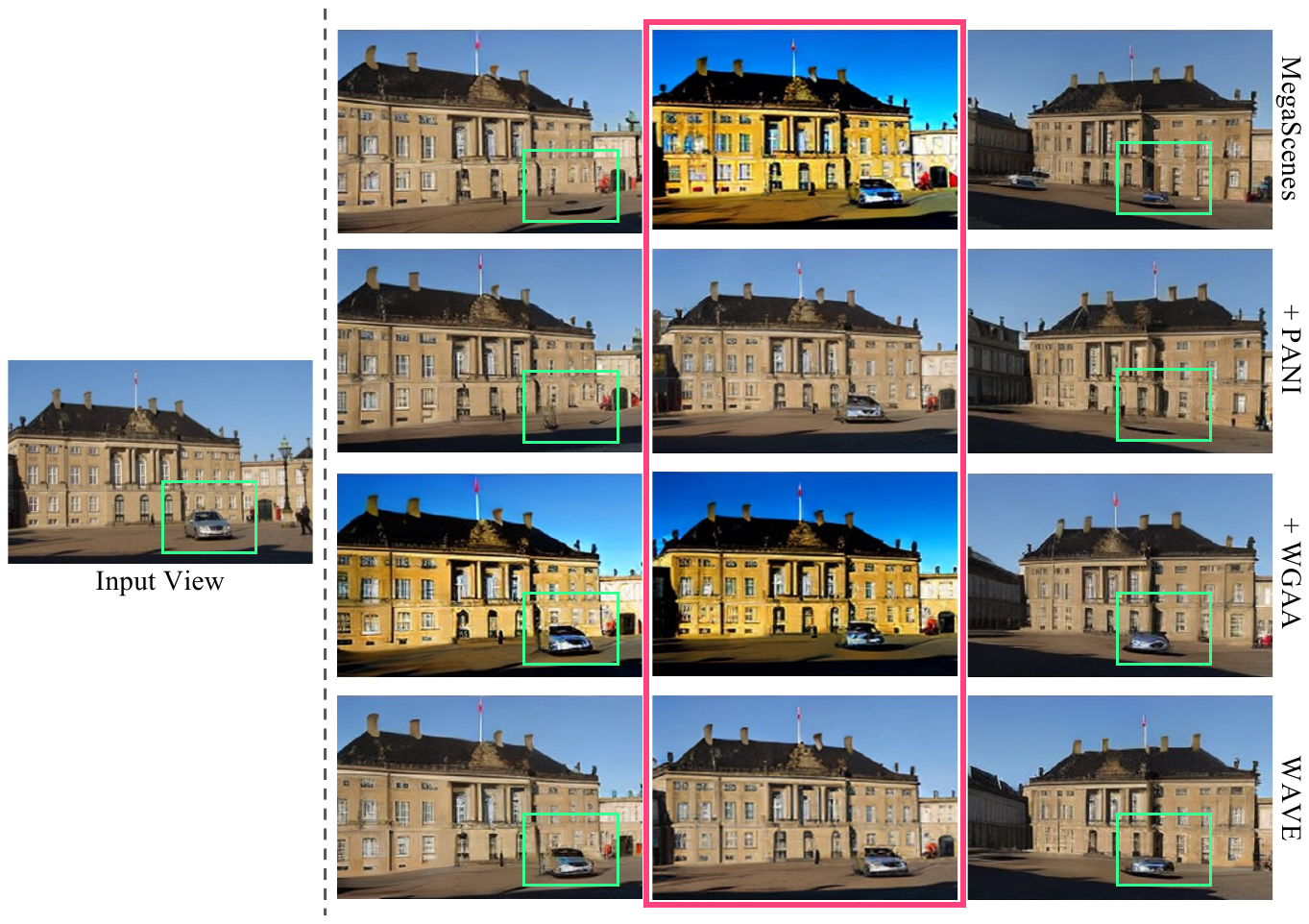} 
    \vspace{-18pt}
    \caption{\textbf{Visual comparisons in ablation study.} We compare of the generated results when PANI and WGAA are removed from our method, respectively. We generate images along an orbital trajectory, ranging from a specific leftward angle to a specific rightward angle relative to the input image.}
    \vspace{-13pt}
    \label{fig:ablation_qual}
\end{figure}

We conduct ablative experiments to study the effect of each component in our method. Ablation experiments are performed on the MegaScenes dataset and MegaScenes pre-trained model \cite{megascenes}. The quantitative and qualitative results are shown in \nametab{}~\ref{tab:ablation} and \namefig{}~\ref{fig:ablation_qual}, respectively.

\vspace{2pt}
\noindent\textbf{Effect of warp-guided adaptive attention.} 
Warp-guided adaptive attention improves performances across all metrics compared to the base model in \nametab{}~\ref{tab:ablation}. 
As shown in \namefig{}~\ref{fig:ablation_qual}, it helps maintain the spatial structure of most scenes, although color variations persist. This indicates that our attention mechanism preserves structural features by sharing information from other generative processes.
Additionally, a detailed comparison with conventional batch self-attention is provided in the supplementary material.
\begin{table}[t!]
\centering
\renewcommand{\arraystretch}{1.2} 
\caption{\textbf{Ablation results.} 
We analyze different components of our model to assess their impact on the overall performance. PANI refers to the pose-aware noise initialization in \namesec{}~\ref{sec:warp_init}, while WGAA denotes the warp-guided adaptive attention in \namesec{}~\ref{sec:adaptive}. }
\vspace{-5pt}
\resizebox{\linewidth}{!}{%
    \begin{tabular}{cccccccc}
    \specialrule{1pt}{0pt}{0pt}
    {\footnotesize Exp.} & \multirow{2}{*}{{PANI}} & \multirow{2}{*}{{WGAA}} & \multicolumn{2}{c}{{\textit{First}}} & \multicolumn{2}{c}{{\textit{Next}}} & {Angular} \\
    \cline{4-5} \cline{6-7} 
    \# & & & {LPIPS $\downarrow$} & {CLIPSIM $\uparrow$} & {LPIPS $\downarrow$} & {CLIPSIM $\uparrow$} & {Consistency $\downarrow$} \\
    \specialrule{0.7pt}{0pt}{0pt}
    1 & & & 0.609    & 0.813    & 0.607     & 0.861    & 20.35    \\
    2 & \checkmark & & 0.584 & 0.819 & 0.560 & 0.873 & 19.76      \\
    3 & & \checkmark & 0.581 & 0.822 & 0.416 & 0.918 & 16.49     \\
     \rowcolor[gray]{0.95} 4 & \checkmark & \checkmark & \textbf{0.561}           & \textbf{0.826}           & \textbf{0.397}           & \textbf{0.920}           & \textbf{15.91}       \\
    \specialrule{1pt}{0pt}{0pt}
    \end{tabular}
    
}
\vspace{-4pt}

\label{tab:ablation}
\end{table}

\vspace{2pt}
\noindent\textbf{Effect of pose-aware noise initialization.}
Applying pose-aware noise initialization also improves performance over the diffusion model in \nametab{}~\ref{tab:ablation}.  
This shows that pose-aware noise initialization effectively reflects view transformations, helping to reduce noise randomness.
As shown in \namefig{}~\ref{fig:ablation_qual}, while objects fail to appear across generated views, color variations do not occur. These results imply that the randomness in noise contributes to color inconsistencies.

\subsection{Downstream Task} \label{down}
Additionally, we validate that the consistent images generated using our method effectively enhance the performance of 3D rendering tasks on various datasets \cite{megascenes, dtu, re10k, mip}. Since 3D rendering tasks rely on multiple images, ensuring view consistency is crucial for providing high-quality multi-view information to 3D models, ultimately improving 3D rendering performance \cite{Mvdreamer, wonder3d}. Thus, this task can be used to evaluate whether the models generate consistent images.
Previous studies \cite{ZeroNVS,dreamgaussian,let2d} on diffusion-based 3D rendering with a single image have explored training 3D models using novel view synthesis diffusion model's guidance. However, this approach makes it challenging to assess the diffusion model itself, as it jointly optimizes the 3D model’s loss and diffusion guidance loss. 

We conduct an evaluation by directly using the generated outputs from the diffusion model as inputs to a 3D model to assess rendering performance. 
In this experiment, VistaDream is excluded since its pipeline already includes 3D rendering.
3D rendering experiments are conducted using 3D Gaussian Splatting \cite{3dgs} with 19 image sets generated by our method and other baseline models \cite{ZeroNVS, megascenes}. As shown in \nametab{}~\ref{tab:3dgs}, our approach outperforms the original methods. These results indicate that our generated consistent images enhance geometric accuracy and visual coherence in 3D reconstruction and rendering tasks. The potential for future application of our method in diffusion-based 3D rendering tasks is also demonstrated.  
Results for datasets not included in \nametab{}~\ref{tab:3dgs} are in the supplementary material.
\begin{table}[t!]
\centering
\setlength{\tabcolsep}{11pt}
\renewcommand{\arraystretch}{1.1} 
\caption{\textbf{3D rendering downstream task.} We evaluate 3D rendering performance using 3D Gaussian Splatting \cite{3dgs} with images generated by our method and baselines. In Mip-NeRF 360, ZeroNVS \cite{ZeroNVS} fails to generate reliable images without a rendering technique, making it unsuitable for 3D rendering.}
\vspace{-5pt}

\resizebox{\linewidth}{!}{%
    \begin{tabular}{p{4cm} ccc}
    
    \specialrule{0.9pt}{0pt}{0pt}
    
    {Method} & {PSNR $\uparrow$} & {SSIM $\uparrow$} & {LPIPS $\downarrow$} \\
    
    \specialrule{0.7pt}{0pt}{0pt}
    
    \multicolumn{4}{l}{\textbf{\textit{MegaScenes}}} \\
    ZeroNVS          & 23.38       & \textbf{0.869}      & 0.098       \\
    \rowcolor[gray]{0.95} ZeroNVS + \ourmodels{}   & \textbf{24.23}       & 0.852       & \textbf{0.094}       \\
    MegaScenes & 22.11       & 0.798       & 0.128       \\
    \rowcolor[gray]{0.95} MegaScenes + \ourmodels{}        & \textbf{24.69}       & \textbf{0.836}       & \textbf{0.096}       \\
    
    \specialrule{0.4pt}{0pt}{0pt}
    
    \multicolumn{4}{l}{\textbf{\textit{DTU}}} \\
    ZeroNVS          & 23.53 & 0.847   & 0.127  \\
    \rowcolor[gray]{0.95} ZeroNVS + \ourmodels{}   & \textbf{25.56} & \textbf{0.871}  & \textbf{0.101}  \\
    MegaScenes & 20.43 & 0.809  & 0.142  \\
    \rowcolor[gray]{0.95} MegaScenes + \ourmodels{}        & \textbf{26.95} & \textbf{0.895}  & \textbf{0.078}  \\
    
    \specialrule{0.4pt}{0pt}{0pt}
    
    \multicolumn{4}{l}{\textbf{\textit{Mip-NeRF 360}}} \\
    ZeroNVS          & - & -  & -  \\
    \rowcolor[gray]{0.95} ZeroNVS + \ourmodels{}   & - & -  & -  \\
    MegaScenes & 25.07 & 0.829   & \textbf{0.074}  \\
    \rowcolor[gray]{0.95} MegaScenes + \ourmodels{}        & \textbf{26.30} & \textbf{0.839}  & 0.082  \\
    
    \specialrule{0.9pt}{0pt}{0pt}
  
    \end{tabular}
}

\vspace{-10pt}
\label{tab:3dgs}
\end{table}

\let\thefootnote\relax\footnote{\scriptsize{{\bf Acknowledgement.} 
 This work was supported in part by IITP RS-2024-00457882 (National AI Research Hub Project), IITP 2020-0-01361, NRF RS-2024-00345806,  NRF RS-2023-00219019, G21004687962, and RS-2024-00403860 (Korea Basic Science Institute, National Research Facilities and Equipment Center). 
}}

\section{Conclusion}
\label{section:conclusion}

We propose a training-free method to improve view consistency in diffusion models for scene-level novel view synthesis from a single image, offering a practical solution for this task.
By leveraging inter-view information through attention and noise reinitialization with warped images, our approach overcomes key limitations of diffusion models, including independent generation and noise randomness.
Lastly, we address the lack of reliable evaluation metrics by introducing a comprehensive metric process, providing a valuable benchmark for future research.

\vspace{3pt}
\noindent\textbf{Limitations and future work.}
The use of warped images makes the method sensitive to large viewpoint changes, leading to significant missing regions.
An autoregressive approach, generating images within a specific view range and iteratively using them for subsequent generations, could address this issue through gradual warping and offer a promising direction for future work.


{
    \small
    \bibliographystyle{ieeenat_fullname}
    \bibliography{main}
}
\clearpage

\onecolumn
\appendix
\section{Code and Website}
A project website has been created to introduce WAVE in a simple and accessible manner, as well as to showcase a variety of qualitative results, implementation codes and videos. It can be accessed via the following link: {\textit{project page: \url{https://jwoo-park0.github.io/wave.github.io/}}}
\section{More Details of Methods} \label{section_supp:aaa}

\subsection{Validating the assumption of WGAA}

We propose \textbf{warp-guided adaptive attention}, WGAA, to manipulate attention, utilizing warped region masks.
This approach is based on the hypothesis that the attention mechanism in the U-Net decoder layers preserves spatial position correspondence. 
Leveraging this assumption, we design warped region masks that align with the attention maps, enabling direct attention manipulation. 
This method efficiently incorporates the novel viewpoint information into the attention mechanism, contributing to generating consistent images.
\begin{figure}[b!]
    \centering
    \includegraphics[width=\linewidth]{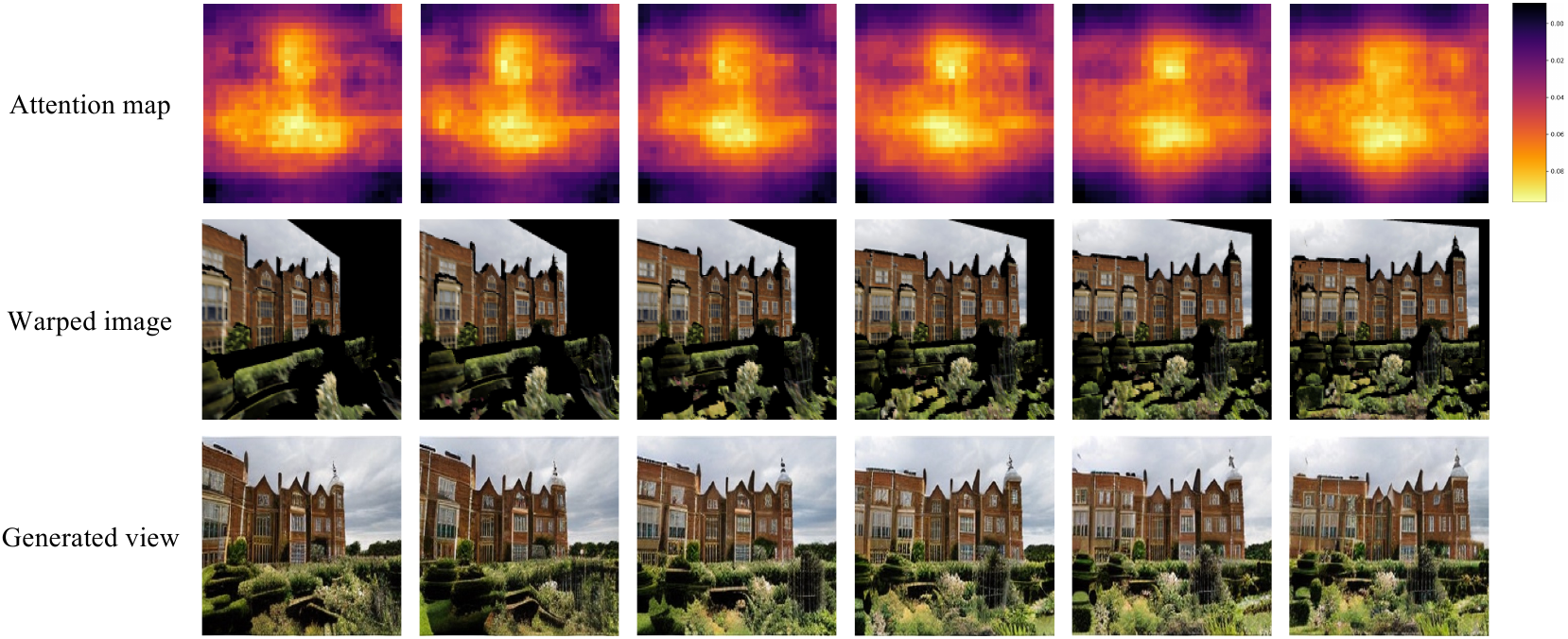} 
    \caption{\textbf{Visualization of attention maps, warp images and generated views.} We present the attention maps of the model alongside the warp images to validate the assumptions of our proposed warp-guided adaptive attention. The results demonstrate that the attention map adapts in a similar manner to changes in the warp image.}
    \label{fig:supple_1-2} 
\end{figure}

To validate the hypothesis above, attention maps are extracted from the U-Net decoder layers. 
Attention maps are extracted at every DDIM \cite{ddim} step, with those at different resolutions interpolated and averaged.
Specifically, attention maps are examined to analyze how the attention distribution varies across different viewpoints. 
As demonstrated in Fig.~\ref{fig:supple_1-2}, the attention maps dynamically adjust based on the warped images. 
This observation confirms our initial hypothesis that the decoder attention retains spatial position correspondence. Thus, our study substantiates the proposed approach.

\section{Metrics}
\label{section_supp:bbb}

\subsection{Video consistency metrics} \label{supple:consistency}
We introduce \textbf{LPIPS-\textit{first}}, \textbf{CLIPSIM-\textit{first}}, \textbf{LPIPS-\textit{next}}, and \textbf{CLIPSIM-\textit{next}}, the metrics proposed in previous works \cite{storydiffusion, sparsectrl} to evaluate video consistency. Originally designed to assess frame-to-frame consistency in video synthesis, these metrics are similarly employed to measure view consistency across generated images under different viewpoints in this work. 

\begin{figure}[t!]
    \centering
    \includegraphics[width=0.9\linewidth]{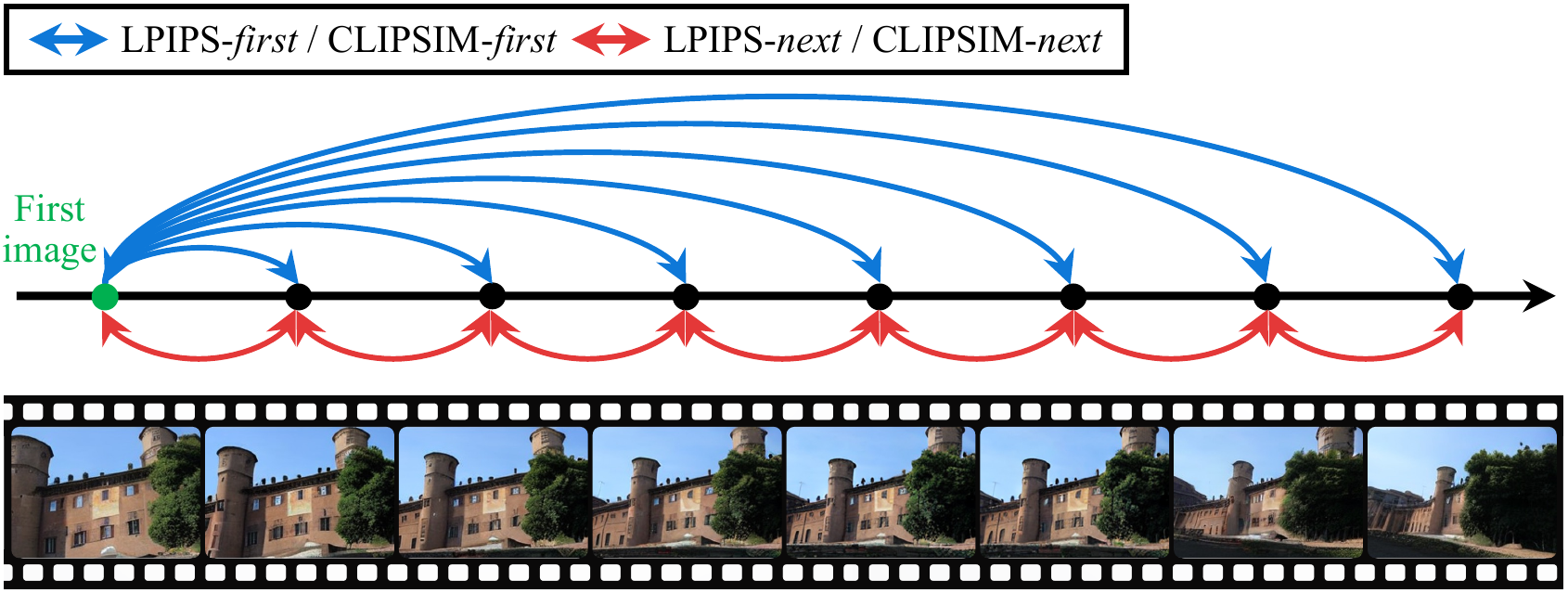}
    \caption{\textbf{Illustration of consistency metrics.} A figure that illustrates the primary consistency metrics used in this study: LPIPS-\textit{first}, CLIPSIM-\textit{first}, and LPIPS-\textit{next}, CLIPSIM-\textit{next} \cite{storydiffusion, sparsectrl}. The \textit{first} metric measures LPIPS and CLIP similarity by comparing images from different viewpoints against the input viewpoint's image. In contrast, the \textit{next} metric computes LPIPS and CLIP similarity by comparing images from adjacent viewpoints.}
    \label{fig:metric}
\end{figure}

First of all, LPIPS \cite{lpips} is a widely used metric for evaluating the perceptual similarity between two images. 
It computes similarity by passing each image through a VGG network \cite{vgg}, extracting feature representations from intermediate layers, and comparing them. Mathematically, it is expressed as follows:
\begin{equation}
    \text{LPIPS}(I_1, I_2) = \sum_l w_l \left\| \phi_l(I_1) - \phi_l(I_2) \right\|_2^2
\end{equation}
where:  \( I_1, I_2 \) are the input images, \( \phi_l(I) \) denotes the deep feature representation from layer \( l \) of a pre-trained network, \( w_l \) is a learned weight for layer \( l \), \( \|\cdot\|_2 \) represents the Euclidean norm. And, CLIP Similarity \cite{clipscore} is a metric that measures the similarity between two images using the CLIP model. Specifically, it evaluates how similar the output embeddings are after passing the images through CLIP’s encoder.
\begin{equation}
    \text{CLIPScore}(I_1, I_2) = \frac{\psi(I_1) \cdot \psi(I_2)}{\|\psi(I_1)\| \|\psi(I_2)\|}
\end{equation}
where: \( \psi(I) \) represents the CLIP embedding of image \( I \), \( \cdot \) denotes the dot product, \( \|\cdot\| \) represents the Euclidean norm.
As illustrated in Fig.~\ref{fig:metric}, LPIPS-\textit{first} and CLIPSIM-\textit{first} calculate LPIPS and CLIP similarity between the input viewpoint and all other viewpoints. 
In contrast, LPIPS-\textit{next} and CLIPSIM-\textit{next} measure consistency between adjacent viewpoints. 
Since the Next metrics are less affected by viewpoint changes compared to the First metrics, they provide a more effective measure of view consistency.

\vspace{3mm}

\subsection{Camera parameter accuracy}
View consistency is evaluated using the metrics outlined in \namesec{}~\ref{supple:consistency}. However, these metrics are inherently limited in providing a comprehensive assessment. 
Consequently, additional evaluation is required for a more thorough analysis.
Specifically, we propose estimating the camera viewpoint of the generated novel view synthesis images and comparing it against the ground truth camera parameters. This approach captures consistency aspects that previous metrics fail to address. 
To quantify camera parameter accuracy, three evaluation metrics are employed: (1) Frobenius Norm, (2) Rotation Angle Difference, and (3) Angular Consistency.

\vspace{3pt}
\noindent\textbf{Frobenius Norm} Frobenius Norm is a matrix norm that extends the Euclidean norm to matrices. It is defined as the square root of the sum of the absolute squares of the elements of a matrix. This norm provides a measure of the overall size of a matrix. Since the camera extrinsic parameter is a matrix composed of 3x4 or 4x4, it could provide the camera parameter accuracy between the estimated camera parameter and the ground truth camera parameter. Given the difference $|a_{ij}|$ between the measured camera parameters and the ground-truth camera parameters, the Frobenius Norm  $\| A \|_F$ is calculated as:
\begin{equation}
\| A \|_F = \sqrt{\sum_{i=1}^{m} \sum_{j=1}^{n} |a_{ij}|^2}.
\end{equation}

\vspace{3pt}
\noindent\textbf{Rotation Angle Difference}
Rotation Angle Difference measures the angular discrepancy between two rotation matrices or quaternions. It is commonly used in 3D vision, robotics, and computer graphics to quantify rotational errors. Given two rotation matrices \( R_1, R_2 \in SO(3) \), the Rotation Angle Difference \( \theta \) is computed as:
\begin{equation}
\theta = \cos^{-1} \left( \frac{\text{trace}(R_1^T R_2) - 1}{2} \right).
\end{equation}

\vspace{3pt}
\noindent\textbf{Angular Consistency}
Angular Consistency refers to the property that ensures the relative orientation between viewpoints remains stable across transformations. It is particularly important in applications such as novel view synthesis, 3D reconstruction, and camera pose estimation. To quantify angular consistency, given a set of rotation matrices \( R_i \) associated with different viewpoints, the angular deviation between consecutive viewpoints can be measured as:
\begin{equation}
\theta_i = \cos^{-1} \left( \frac{\text{trace}(R_i^T R_{i+1}) - 1}{2} \right).
\end{equation}
For a sequence of rotations, the overall angular consistency error can be defined as the variance of the angular differences where \( \bar{\theta} \) is the mean rotation angle difference:
\begin{equation}
E_{\text{angular}} = \frac{1}{N-1} \sum_{i=1}^{N-1} (\theta_i - \bar{\theta})^2 , \  \  \bar{\theta} = \frac{1}{N-1} \sum_{i=1}^{N-1} \theta_i.
\end{equation}
Ensuring low angular consistency error is critical for maintaining coherence in generated views and preventing distortions in viewpoint transitions.

\section{Experiment Details}
\label{section_supp:ccc}

This section outlines the experimental setup, and additional experiments. We begin by describing the implementation details (\namesec{}~\ref{implement}) and then introduce supplementary experiments conducted to further validate the proposed method. 
The experiments are divided into four main parts. 

First, the primary experiment is described, focusing on the evaluation of the consistency metric and camera accuracy, as detailed in \namesec{}~\ref{main exp}. 
Next, the target-view image reconstruction metric experiments conducted on the MegaScenes dataset \cite{megascenes}, where images are generated for specific viewpoints rather than continuous ones, are presented in \namesec{}~\ref{recon}. 
We then detail image reconstruction experiments on the RealEstate10K(RE10K) sequence dataset in \namesec{}~\ref{re10k sqe}. Finally, the methodology for extracting camera poses and performing the 3D Gaussian Splatting rendering task is explained in \namesec{}~\ref{3dgs,camerapose}.

\subsection{Implementation details} \label{implement}
For the novel view synthesis diffusion model, this work utilizes pre-trained models such as ZeroNVS \cite{ZeroNVS} and MegaScenes \cite{megascenes}, which generate images at a resolution of 256×256, maintained across all experiments. DDIM with 50 inference steps is adopted for sampling. Additionally, the attention map dropout technique from \citet{consistory} is implemented, using a fixed dropout ratio of 0.2, with no significant performance changes observed when varying the dropout ratio. For noise initialization, a Gaussian low-pass filter is applied, following the filter used in prior studies \cite{freeinit,consisti2v}. During the noise application to latent variables, the noise level is fixed at 950 after the latent variables pass through the VAE encoder. The warp algorithm is implemented using the Pyrender library, equal to MegaScenes. Camera parameters are also configured in the OpenGL environment, as required by Pyrender. For warping, the depth map is extracted using the Depth Anything model \cite{depth}, a monocular depth estimation model, following the approach of the previous work, MegaScenes. All experiments are conducted using an RTX A6000 GPU.

\subsection{Consistency and camera accuracy experiment} \label{main exp}
The experiments are conducted using the MegaScenes, RE10K, and DTU datasets. Mip-NeRF 360 dataset is excluded due to its limited number of scenes. The evaluation focuses on assessing image viewpoint consistency across consecutive viewpoints. These viewpoints are arranged in an \textit{orbit pose}, with the input image set as the central viewpoint and surrounding poses forming an orbit configuration. The variation in \textit{orbit pose} is illustrated as a gray line in \namefig{}~\ref{fig:camera}, and in this study, we define the orbit pose with 19 viewpoints, fixing the radius at 1 and setting the rotation angle to 30 degrees.
Thus, this experiment evaluates how well models generate consistent images when given a single input image and an orbit pose as input.

Camera parameters are generally divided into intrinsic and extrinsic parameters, with extrinsic parameters primarily used for camera pose evaluation. 
Extrinsic parameters consist of a rotation matrix and a translation matrix. However, since the translation matrix is highly sensitive to scale, our evaluation process focuses on the rotation matrix.
To evaluate camera accuracy, the generated images are saved from the previous experiment and COLMAP \cite{colmap1,colmap2} is used to extract camera poses. COLMAP is chosen because it remains widely used for constructing camera parameter datasets \cite{co3d,instant}. Since images are generated using an orbit pose, we use the corresponding ground truth camera parameters for evaluation. 

While most cases in camera parameter estimation are suitable for evaluation, some instances result in failure cases. During the camera parameter estimation process for image sets, there are cases where the expected 19 camera parameters in the orbit pose setup are not obtained. This issue arises when COLMAP's SIFT algorithm fails to find correspondences between images, leading to missing camera parameters. Since the inability to establish correspondences indicates a lack of image consistency, we apply a strong penalty when evaluating camera accuracy.
For example, when comparing the estimated camera parameters to the ground-truth orbit pose, cases, where only 2 out of 19 cameras are reconstructed, are handled by duplicating the available parameters to match the required 19 viewpoints before evaluation. This ensures that inconsistent images, which fail to establish correspondences, receive a penalty. 
However, simply duplicating poses in this manner may not be suitable for all scenarios, as the criteria for duplication vary depending on the number of available poses. It would be possible to apply the maximum error to camera poses that are not successfully estimated.
For Frobenius Norm, the maximum error can be defined by comparing it with the identity matrix. Similarly, for Rotation Angle Difference and Angular Consistency, the maximum error is typically set to $\pi$ (180° radians). Leveraging these predefined values, we can introduce penalty-based adjustments, which can be further explored in future evaluations.

\subsection{Target view image reconstruction experiment} \label{recon}
The target view image reconstruction experiment differs from the evaluation of consistency between consecutive images.
This experiment involves generating paired images for different viewpoints, where the model generates a corresponding image from another viewpoint given an image from one viewpoint. Unlike the previous experiment, which measures viewpoint consistency, this experiment evaluates the model's ability to accurately generate images for specific viewpoints. We conduct this experiment to evaluate whether our method negatively impacts performance in this aspect. This experiment follows a similar approach to MegaScenes \cite{megascenes}. 

\begin{table*}[t!]
\centering
\setlength{\tabcolsep}{9pt}
\caption{\textbf{Target view image generation evaluation.} Quantitative evaluation of our method across diverse datasets (MegaScenes, DTU, RE10K, and Mip-NeRF 360). 
We report PSNR, SSIM, and LPIPS to evaluate reconstruction quality. To assess the consistency of the warped image regions, we include Masked PSNR, Masked SSIM, and Masked LPIPS \cite{megascenes}. Additionally, FID and KID are reported to measure the overall quality of image generation. Our approach improves performance over the baseline methods (MS, ZeroNVS) across overall metrics, demonstrating better view consistency and reconstruction fidelity.
}
\resizebox{\textwidth}{!}{%
    \begin{tabular}{p{3.5cm} cccccccc}
    \toprule
    {Method} & {PSNR $\uparrow$} & {SSIM $\uparrow$} & {LPIPS $\downarrow$} & {Masked PSNR $\uparrow$} & {Masked SSIM $\uparrow$} & {Masked LPIPS $\downarrow$} & {FID $\downarrow$} & {KID $\downarrow$} \\
    \midrule
    \multicolumn{9}{l}{\textbf{\textit{MegaScenes}}} \\
    ZeroNVS          & 7.69 & 0.150 & 0.611 & 11.09 & 0.653 & 0.268             & 46.05 & 0.029 \\
    \rowcolor[gray]{0.95} ZeroNVS + \ourmodels{}   & \textbf{10.70}            & \textbf{0.335}            & \textbf{0.497}            & \textbf{18.14}           & \textbf{0.823}           & \textbf{0.160}             & \textbf{32.93}           & \textbf{0.016}             \\
    MegaScenes & 12.28 & 0.432 & 0.395 & 23.36 & 0.877 & 0.094 & \textbf{13.55} & 0.005 \\
    \rowcolor[gray]{0.95} MegaScenes + \ourmodels{}        & \textbf{12.49} & \textbf{0.438} & \textbf{0.392} & \textbf{24.496} & \textbf{0.878} & \textbf{0.092} & 15.03 & \textbf{0.005} \\
    \midrule
    \multicolumn{9}{l}{\textbf{\textit{DTU}}} \\
    ZeroNVS          & 8.94    & 0.213   & 0.644   &  18.05  & 0.627   & 0.286             & 63.27   & 0.016      \\
    \rowcolor[gray]{0.95} ZeroNVS + \ourmodels{}   & \textbf{11.11}  & \textbf{0.347}   & \textbf{0.587}  &  \textbf{19.81}  & \textbf{0.706} & \textbf{0.246}             & \textbf{57.65} & \textbf{0.008}    \\
    MegaScenes & 10.88  & \textbf{0.394} & 0.503 & 20.36 & 0.722 & \textbf{0.211}             & 34.96   & 0.007       \\
    \rowcolor[gray]{0.95} MegaScenes + \ourmodels{}        & \textbf{11.95}  & 0.386 & \textbf{0.460} & \textbf{20.72} & \textbf{0.722} & 0.212            & \textbf{22.59} & \textbf{0.004} \\
    \midrule
    \multicolumn{9}{l}{\textbf{\textit{RE10K}}} \\
    ZeroNVS          & 12.27    & 0.263   & 0.520  & 20.25 & 0.653   & 0.251              & 8.05    & 0.002     \\
    \rowcolor[gray]{0.95} ZeroNVS + \ourmodels{}   & \textbf{12.33}            & \textbf{0.269}            & \textbf{0.514}            & \textbf{20.33}           & \textbf{0.657}           & \textbf{0.248}             & \textbf{6.15}           & \textbf{0.001}             \\
    MegaScenes & 11.62  & \textbf{0.309} & \textbf{0.494} & 19.98 & 0.224 & 0.669             & 15.81 & 0.007 \\
    \rowcolor[gray]{0.95} MegaScenes + \ourmodels{}        & \textbf{11.78}  & 0.261  & 0.516 & \textbf{20.04} & \textbf{0.650} & \textbf{0.252}              & \textbf{10.20} & \textbf{0.004} \\
    \midrule
    \multicolumn{9}{l}{\textbf{\textit{Mip-NeRF 360}}} \\
    ZeroNVS          & 11.00      & 0.123    & 0.675   & 23.89 & 0.768   & 0.209             & \textbf{78.64}  & 0.016 \\
    \rowcolor[gray]{0.95} ZeroNVS + \ourmodels{}   & \textbf{11.48}   & \textbf{0.137}      & \textbf{0.664}    & \textbf{24.32} & \textbf{0.779}    & \textbf{0.201}   & 78.93   & \textbf{0.014}  \\
    MegaScenes & 11.90  & \textbf{0.182} & 0.566 & 25.76 & 0.809 & 0.155              & 58.87  & \textbf{0.009} \\
    \rowcolor[gray]{0.95} MegaScenes + \ourmodels{}        & \textbf{12.29} & 0.181 & \textbf{0.560} & \textbf{26.24} &\textbf{0.813} & 0.155              & \textbf{58.48} & 0.011 \\
    \bottomrule
    \end{tabular}
}

\label{tab:recon}

\end{table*}
Reconstruction experiments are conducted on the MegaScenes \cite{megascenes}, RE10K \cite{re10k}, DTU \cite{dtu}, and Mip-NeRF 360 \cite{mip} datasets. 
For the MegaScenes dataset, we follow the publicly available test code to ensure consistency with previous work \cite{megascenes}. For other datasets, custom test configurations are created due to the inaccessibility of the test code. 
Given the model's inherent limitations in generating images with significant viewpoint changes, paired datasets with closer viewpoint pairs are constructed for RE10K, DTU, and Mip-NeRF 360. 
While our primary focus is improving viewpoint consistency, our method also yields improvements in reconstruction and image metrics. Notably, metrics such as Masked PSNR, Masked LPIPS, and Masked SSIM \cite{megascenes} as for evaluating paired image consistency, shows improved performance in our experimental results. 
The results demonstrate that our approach maintains consistency not only between the generated images but also between the input view and the generated images. Additionally, the experiment shows that addressing the inherent factors of the diffusion models leads to improved image quality in generating novel view images from specific viewpoints.

\subsection{RE10K sequence evaluation} \label{re10k sqe}
RE10K \cite{re10k} dataset differs from other datasets in that it is a sequence dataset composed of videos, where each frame contains corresponding image and camera viewpoint information. 
The dataset consists of videos, so evaluation is restricted to predefined viewpoints rather than arbitrary ones. However, since it can provide consecutive viewpoint ground truth images unlike other datasets, we conduct this experiment to validate our method by evaluating the generated images using image generation and reconstruction metrics.
For evaluation, a random input view image is selected from a video sequence, and 10 frames are skipped between successive viewpoints to generate a total of 6 target viewpoints. The generated images are then compared with their corresponding ground truth frames using LPIPS, PSNR, FID, and KID as evaluation metrics.

\subsection{Downstream task} \label{3dgs,camerapose}
Recent studies have increasingly explored the application of novel-view diffusion models to 3D rendering tasks \cite{sv3d, ZeroNVS, vista}, leveraging the generative capabilities of these models in combination with 3D models for rendering purposes. Rather than solely focusing on novel-view synthesis, this work also investigates the generation of consistent images and evaluates their effectiveness in 3D rendering tasks. First, camera poses are extracted from generated image sets using COLMAP \cite{colmap1,colmap2}. The measured poses and generated images are then used as input to the 3D Gaussian splatting model \cite{3dgs} for rendering, with 3,000 training iterations. In addition, to mitigate potential errors in COLMAP's camera parameter estimation, cases where the number of camera poses is significantly lower than the number of images are excluded from evaluation. Due to space limitations in the main paper, results from additional datasets that couldn't be included are provided in \nametab{}~\ref{tab:3dgs_supple}.
\begin{table}[t!]
\centering
\setlength{\tabcolsep}{9pt}
\renewcommand{\arraystretch}{0.95} 
\caption{\textbf{Additional 3D rendering downstream tasks.} We present the additional quantitative results of experiments from different datasets where 3D Gaussian Splatting \cite{3dgs} performs 3D rendering by using those generated by our method and baseline.}
\resizebox{0.5\linewidth}{!}{%
    \begin{tabular}{p{3.5cm} ccc}
    \toprule
    {Method} & {PSNR $\uparrow$} & {SSIM $\uparrow$} & {LPIPS $\downarrow$} \\
    \midrule
    \multicolumn{4}{l}{\textbf{\textit{DTU}}} \\
    ZeroNVS          & 23.53 & 0.847   & 0.127  \\
    \rowcolor[gray]{0.95} ZeroNVS + \ourmodels{}   & \textbf{25.56} & \textbf{0.871}  & \textbf{0.101}  \\
    MegaScenes & 20.43 & 0.809  & 0.142  \\
    \rowcolor[gray]{0.95} MegaScenes + \ourmodels{}        & \textbf{26.95} & \textbf{0.895}  & \textbf{0.078}  \\
    \midrule
    \multicolumn{4}{l}{\textbf{\textit{RE10K}}} \\
    ZeroNVS          & 24.74 & 0.883  & 0.107  \\
    \rowcolor[gray]{0.95} ZeroNVS + \ourmodels{}   & \textbf{26.48} & \textbf{0.900}  & \textbf{0.088}  \\
    MegaScenes & 21.23 & 0.817   & 0.145  \\
    \rowcolor[gray]{0.95} MegaScenes + \ourmodels{}        & \textbf{24.51} & \textbf{0.876}  & \textbf{0.086}  \\
    \bottomrule
    \end{tabular}
}
\label{tab:3dgs_supple}
\end{table}

\begin{table}[h!]
\centering
\setlength{\tabcolsep}{10pt} 
\caption{\textbf{Camera accuracy in warp-guided adaptive attention.} We show quantitative results to evaluate whether warp-guided adaptive attention improves viewpoint accuracy. Our method, warp-guided adaptive attention, achieves higher camera pose accuracy compared to batch self-attention (\textit{conventional\_att}), which has been used in previous studies.}
\resizebox{\linewidth}{!}{%
    \begin{tabular}{lccc}
    \toprule
     & {Frobenius Norm (Rotation) $\downarrow$} & {Rotation Angle Difference $\downarrow$} & {Angular Consistency $\downarrow$} \\
    \midrule
    \multicolumn{4}{l}{\textbf{\textit{MegaScenes}}} \\ 
    MegaScenes + \textit{conventional\_att}            & 0.412       & 0.299       & 17.13       \\
    \rowcolor[gray]{0.95} MegaScenes + \ourmodels{}           & \textbf{0.382}       & \textbf{0.277}       & \textbf{15.91}       \\
    \midrule
    \multicolumn{4}{l}{\textbf{\textit{DTU}}} \\ 
    MegaScenes + \textit{conventional\_att}            & 0.383   & 0.281   & 16.09  \\
    \rowcolor[gray]{0.95} MegaScenes + \ourmodels{} & \textbf{0.155} & \textbf{0.110} & \textbf{6.32}  \\
    \midrule
    \multicolumn{4}{l}{\textbf{\textit{RE10K}}} \\ 
    MegaScenes + \textit{conventional\_att}           & 0.321 & 0.231 & 13.25  \\
    \rowcolor[gray]{0.95} MegaScenes + \ourmodels{} & \textbf{0.149}  & \textbf{0.108} & \textbf{6.20}  \\
    \bottomrule
    \end{tabular}
}
\label{tab:method-eval}
\end{table}
\section{Additional Experiments}
\subsection{Effect of WGAA on Camera Accuracy}
Conventional batch self-attention \cite{free3d,consistory,storydiffusion} aggregates all key-value pairs to compute the final representation. In contrast, our proposed warp-guided adaptive attention selectively determines the relevant viewpoint range required to generate a specific viewpoint by the adaptive warp-range selection. It then aggregates only the necessary key-vale pairs based on viewpoint changes. 
As shown in the main paper, applying the conventional batch self-attention method, widely used in previous studies, results in reduced camera accuracy. However, since ground-truth images are unavailable, evaluating the generated results alone does not provide a fully reliable assessment. To address this, we conduct a quantitative evaluation of camera accuracy and extend our experiments beyond the MegaScenes dataset to DTU and RE10K datasets. As presented in \nametab{}~\ref{tab:method-eval}, our warp-guided adaptive attention significantly outperforms conventional batch self-attention. This result validates our hypothesis that the reference range should dynamically adjust according to viewpoint changes.

\subsection{Analysis by Depth Map Quality}
We conduct additional experiments to investigate its impact on view consistency. As part of this, we evaluate the performance across different Depth Anything~\cite{depth} model sizes, where smaller models show degraded performance (AbsRel). The table indicates that although the depth metric degrades, view consistency shows little variation across different depth estimation models, outperforming base MegaScenes. This demonstrates the robustness of our method.
\begin{table}[h!]
\centering
\renewcommand{\arraystretch}{1.1} 
\caption{\textbf{Performance by Depth Map Quality.} We evaluate the performance of our method when applied to different model sizes of Depth Anything (DA), and report the corresponding quantitative results.}
\resizebox{0.8\linewidth}{!}{%
    \begin{tabular}{l|c|cccc}
    \specialrule{1pt}{0pt}{0pt}
    Method  & AbsRel $\downarrow$  & {\textit{Next}-LPIPS $\downarrow$} & {\textit{Next}-CLIPSIM $\uparrow$} & {\textit{First}-LPIPS $\downarrow$} & {\textit{First}-CLIPSIM $\uparrow$}   \\
    \specialrule{0.7pt}{0pt}{0pt}
    MegaScenes & - & 0.607 & 0.861 & 0.609 & 0.813 \\
    \specialrule{0.7pt}{0pt}{0pt}
    WAVE (DA-S) & 0.053  & 0.415 \textcolor{ForestGreen}{(-0.192)} & 0.926 \textcolor{ForestGreen}{(+0.065)} & 0.568 \textcolor{ForestGreen}{(-0.041)}& 0.796  \textcolor{gray}{(-0.017)}    \\
    WAVE (DA-B)  & 0.046 & 0.399 \textcolor{ForestGreen}{(-0.208)}  & 0.927 \textcolor{ForestGreen}{(+0.066)} & 0.558 \textcolor{ForestGreen}{(-0.051)}   & 0.804  \textcolor{gray}{(-0.009)}       \\
    WAVE (DA-L) & 0.043 & 0.397 \textcolor{ForestGreen}{(-0.210)} & 0.920  \textcolor{ForestGreen}{(+0.059)} & 0.561 \textcolor{ForestGreen}{(-0.048)} & 0.826 \textcolor{ForestGreen}{(+0.013)}     \\
    \specialrule{1pt}{0pt}{0pt}
    \end{tabular}
}
\label{tab:ablation}
\end{table}
\subsection{More Comparisons with PANI}
We conduct an additional experiment to evaluate PANI by comparing it with other training-free noise initialization methods designed for consistency. Specifically, we compare our method with FrameInit (3D FFT) from ConsistI2V~\cite{consisti2v}. 
Unfortunately, MultiDiff~\cite{multidiff} is excluded from the comparison due to the unavailability of its code. This experiment is conducted on the MegaScenes dataset. The results show that PANI outperforms in both view consistency and camera pose accuracy. 
Unlike FrameInit method for consistent video, which does not incorporate view information, our method integrates such information, enabling the generation of view-consistent images.
\begin{table}[h!]
\centering
\setlength{\tabcolsep}{9pt} 
\renewcommand{\arraystretch}{1.1}
\caption{\textbf{Comparison with other noise initialization method.} Additional experiments are conducted to validate the effectiveness of PANI, comparing it with other noise initialization methods proposed in previous studies. Evaluation is based on both consistency and camera pose accuracy.}
\resizebox{0.8\linewidth}{!}{%
    \begin{tabular}{lccccc}
    \specialrule{1pt}{0pt}{0pt}
    Method  & {\textit{Next}-LPIPS $\downarrow$} & {\textit{Next}-CLIPSIM $\uparrow$} & {\textit{First}-LPIPS $\downarrow$} & {\textit{First}-CLIPSIM $\uparrow$}  \\
    \specialrule{0.7pt}{0pt}{0pt}
    WAVE w/ FrameInit & 0.433 & 0.919 & 0.610 & 0.813      \\
    \rowcolor[gray]{0.95} WAVE w/ PANI   & \textbf{0.397}     & \textbf{0.920}    & \textbf{0.561}    & \textbf{0.826}  \\
    \specialrule{1pt}{0pt}{0pt}
    \end{tabular}
}
\vspace{-5pt}
\label{tab:pani}
\end{table}
\begin{table}[h!]
\centering
\renewcommand{\arraystretch}{1.15} 
\vspace{-20pt}
\resizebox{0.8\linewidth}{!}{%
    \begin{tabular}{lccccc}
    Method  & {Frobenius Norm (Rotation)} & {Rotation Angle Difference} & {Angular Consistency}  \\
    \specialrule{0.7pt}{0pt}{0pt}
    WAVE w/ FrameInit & 0.402 & 0.291 & 16.67      \\
    \rowcolor[gray]{0.95} WAVE w/ PANI & \textbf{0.382}    & \textbf{0.277}    & \textbf{15.91}          \\
    \specialrule{1pt}{0pt}{0pt}
    \end{tabular}
}
\vspace{-12pt}
\label{tab:pani}
\end{table}
\section{More Qualitative Results}
\label{section_supp:ddd}

\subsection{Camera parameter visualization}

\begin{figure*}[h!]
    \centering
    \includegraphics[width=\textwidth]{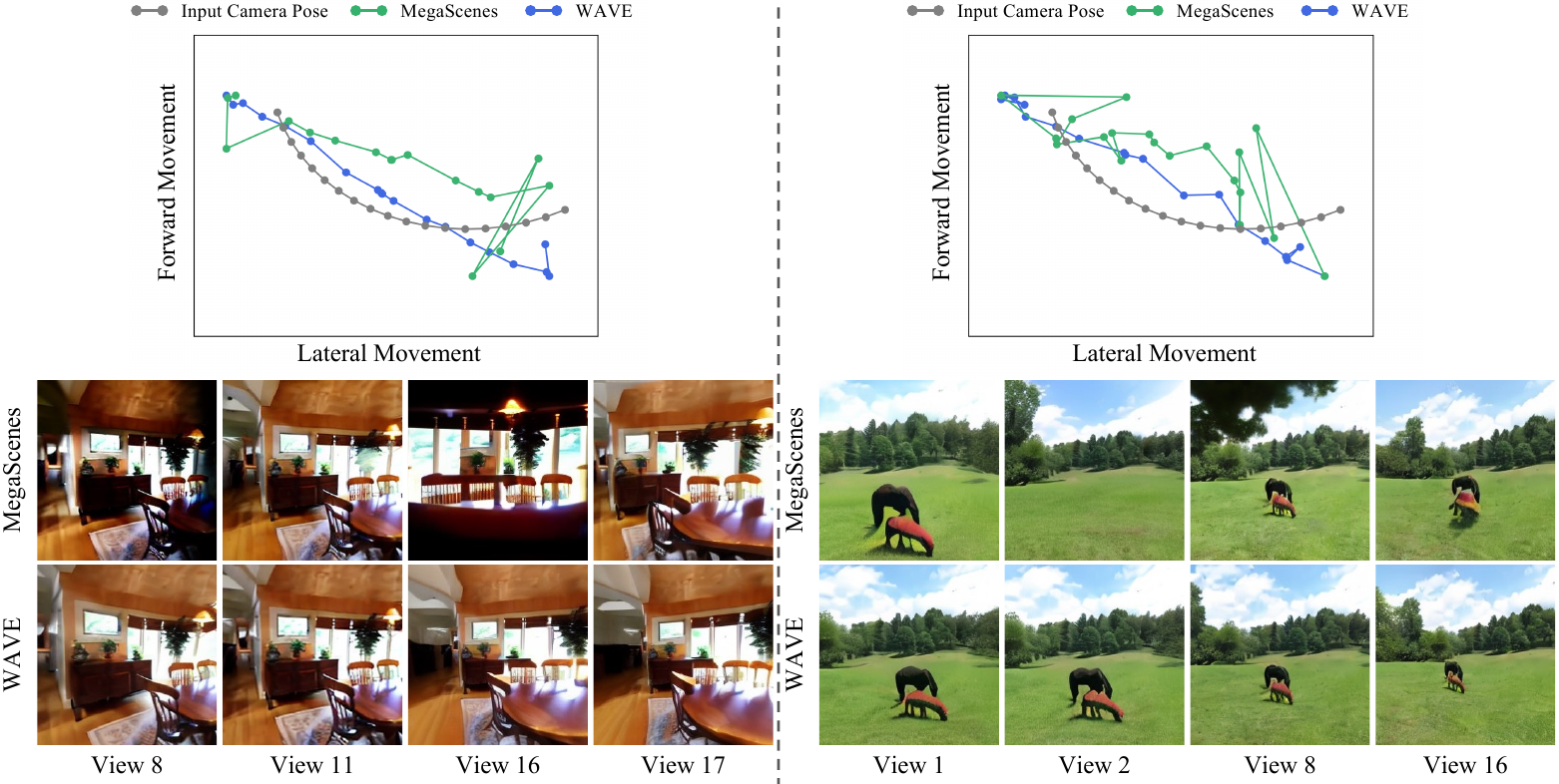}
    \caption{\textbf{Camera parameter visualization.} These examples are taken from the camera accuracy evaluation process. The images below represent the results generated by each model: MegaScenes \cite{megascenes} and MegaScenes + WAVE, while the camera graph above visualizes the measured camera poses from the generated images.} 
    \label{fig:camera}
\end{figure*}

We present a visualization of the camera parameters measured in the camera accuracy experiment. While camera accuracy is evaluated using only the rotation matrix from the camera extrinsic parameters, the translation matrix is visualized to provide further insights into the translation component. 
Since the camera poses are normalized, the height variation is minimal. Therefore, only the x and z coordinates for camera translation are used. To maintain consistency with the ground-truth pose, the measured parameters are also normalized to match the corresponding scale.
As illustrated in the \namefig{}~\ref{fig:camera}, it shows that images with higher view consistency exhibit greater camera accuracy, whereas inconsistent images tend to have lower camera accuracy. This result demonstrates that evaluating camera parameters is an effective approach for assessing view consistency.

\subsection{Additional samples}
\begin{figure*}[b!]
    \centering
    \includegraphics[width=0.95\textwidth]{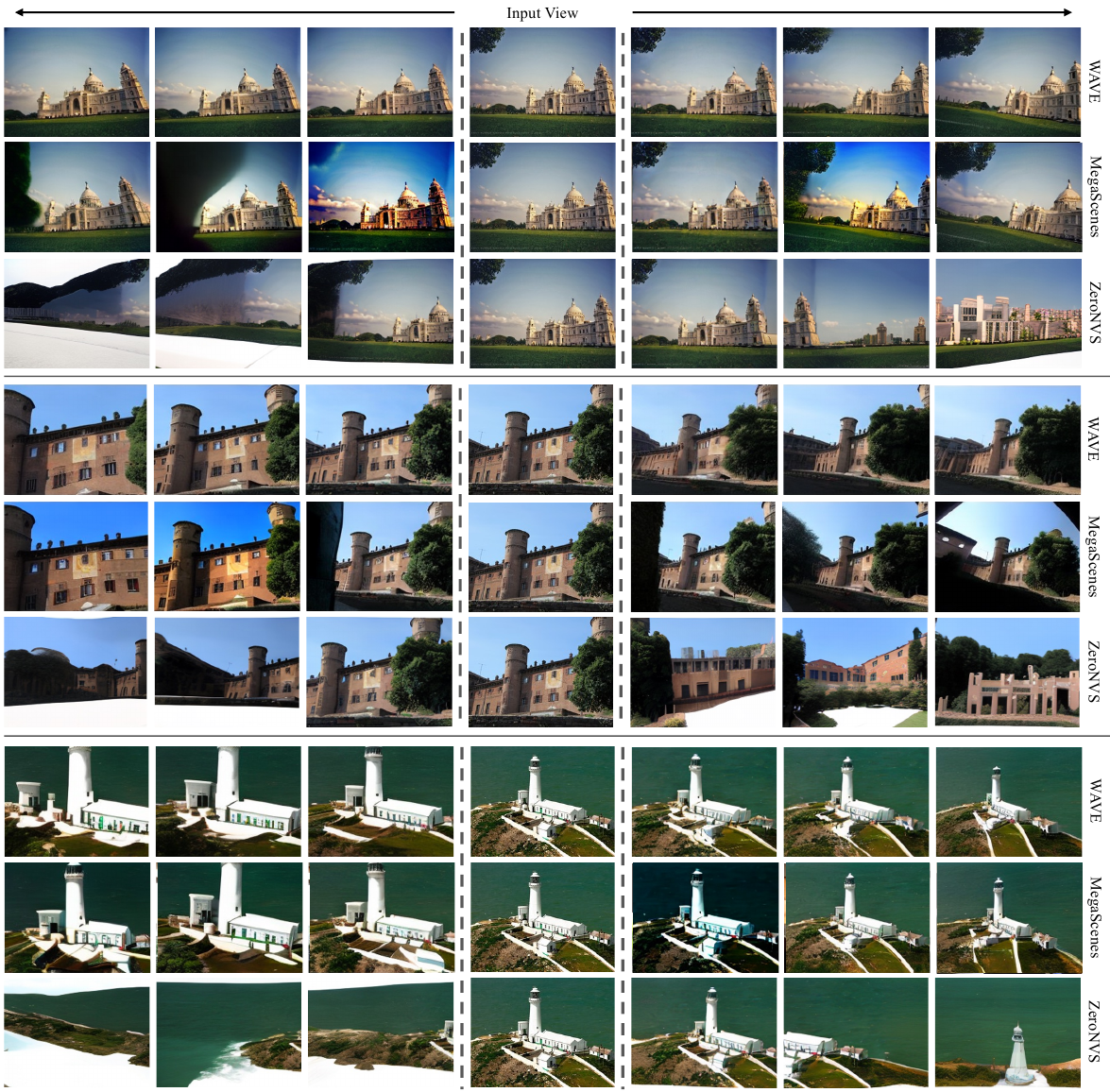} 
    \caption{\textbf{Comparison to diffusion methods on MegaScenes.} We compare our framework with existing diffusion-based models, ZeroNVS \cite{ZeroNVS} and MegaScenes \cite{megascenes}. Additional generation results are provided on the MegaScenes dataset. The images in the middle column represent the input images.}
    \label{fig:supple_2} 
\end{figure*}

\begin{figure*}[t!]
    \centering
    \includegraphics[width=\textwidth]{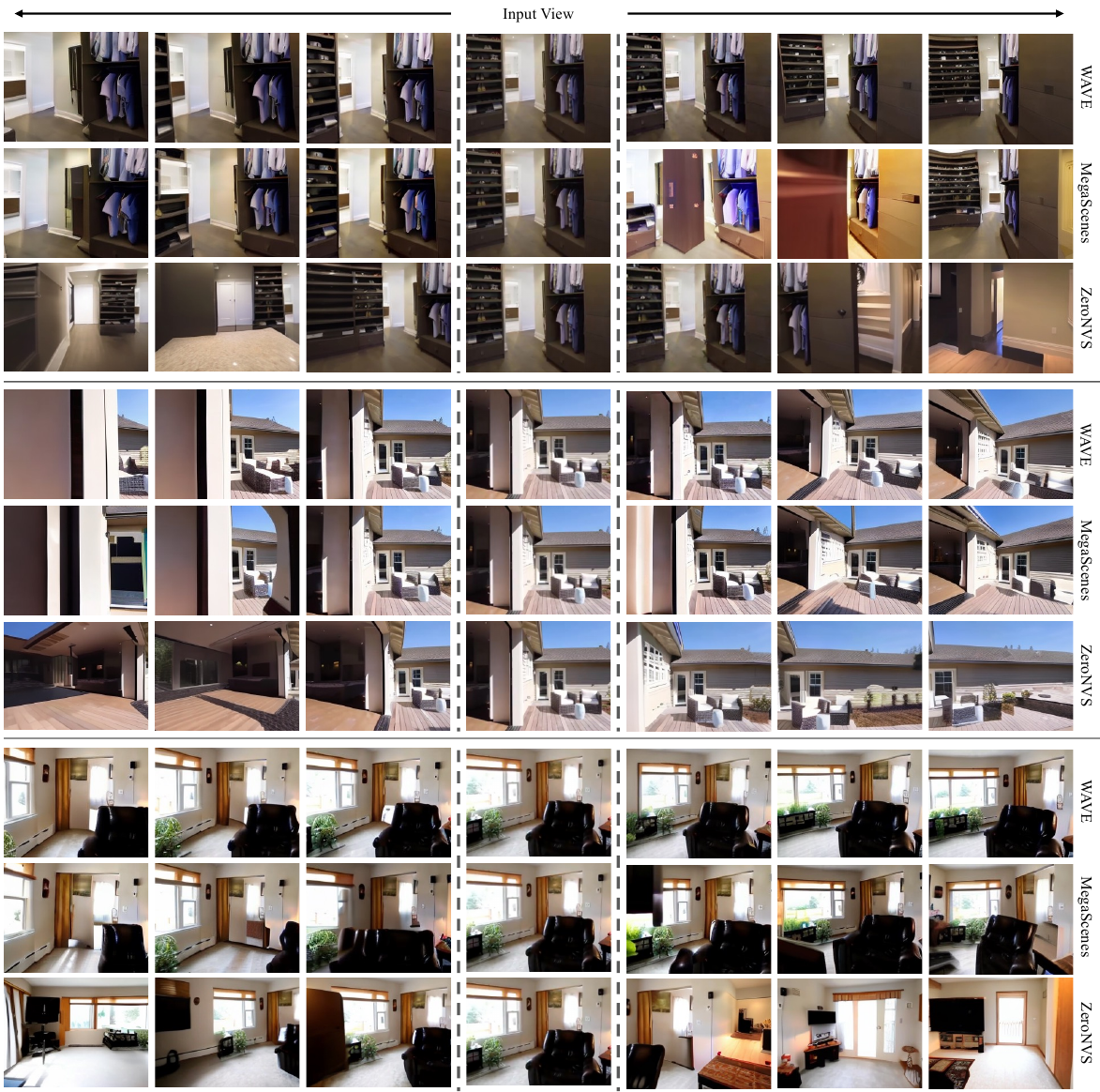} 
    \caption{\textbf{Comparison to diffusion methods on RE10K.} We provide additional generation results on the RE10K \cite{re10k} dataset to compare our method with baselines. The images in the middle column represent the input images.} 
    \label{fig:supple_3} 
\end{figure*}

\begin{figure*}[t!]
    \centering
    \includegraphics[width=0.9\textwidth,height=\textheight,keepaspectratio]{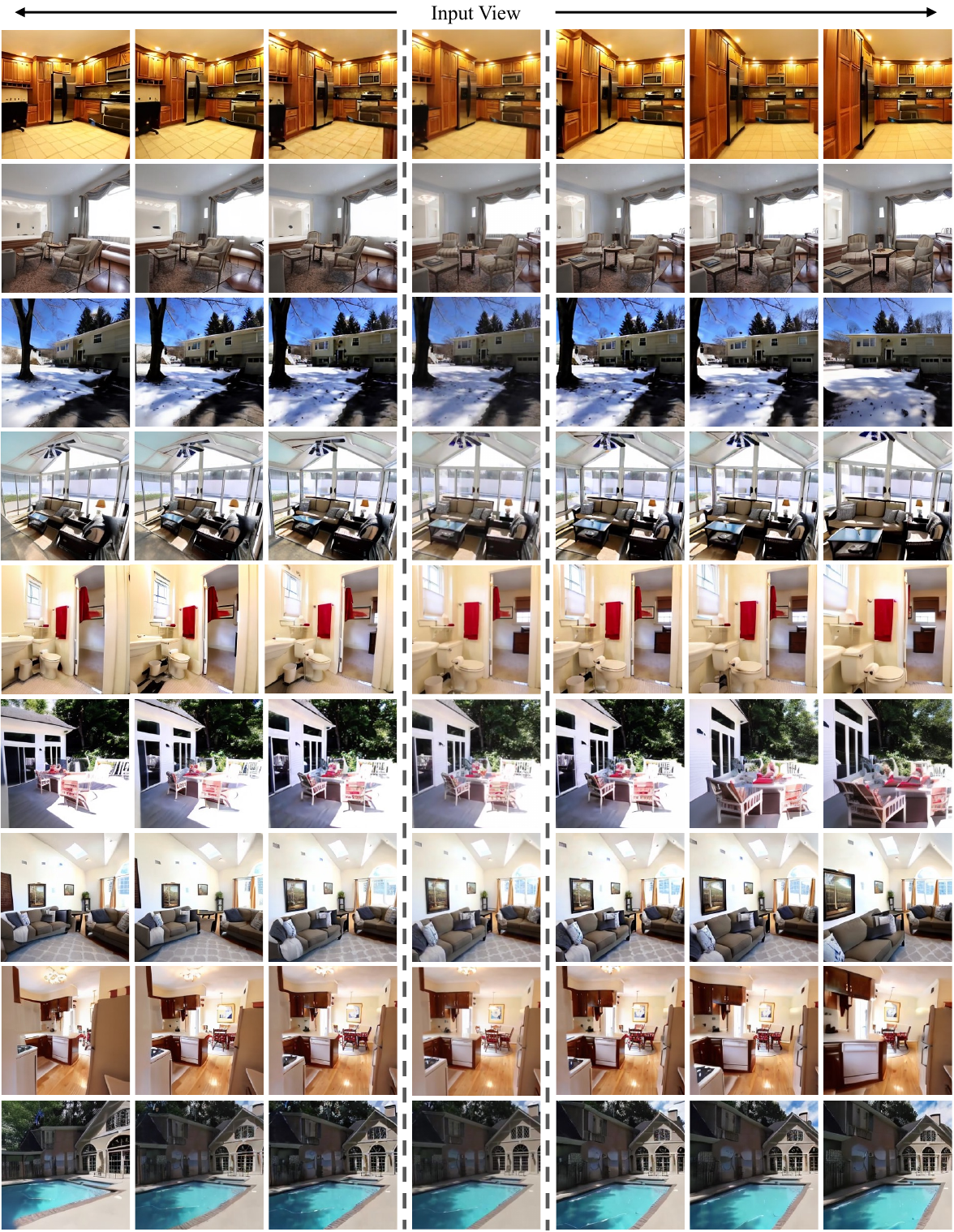} 
    \caption{\textbf{Qualitative results.} Additionally, using samples from the RE10K dataset \cite{re10k}, we generate images with our method by taking the input view image in the middle column and continuous camera poses as inputs. }
    \label{fig:supple_4} 
\end{figure*}

\begin{figure*}[t!]
    \centering
    \includegraphics[width=0.9\textwidth,height=\textheight,keepaspectratio]{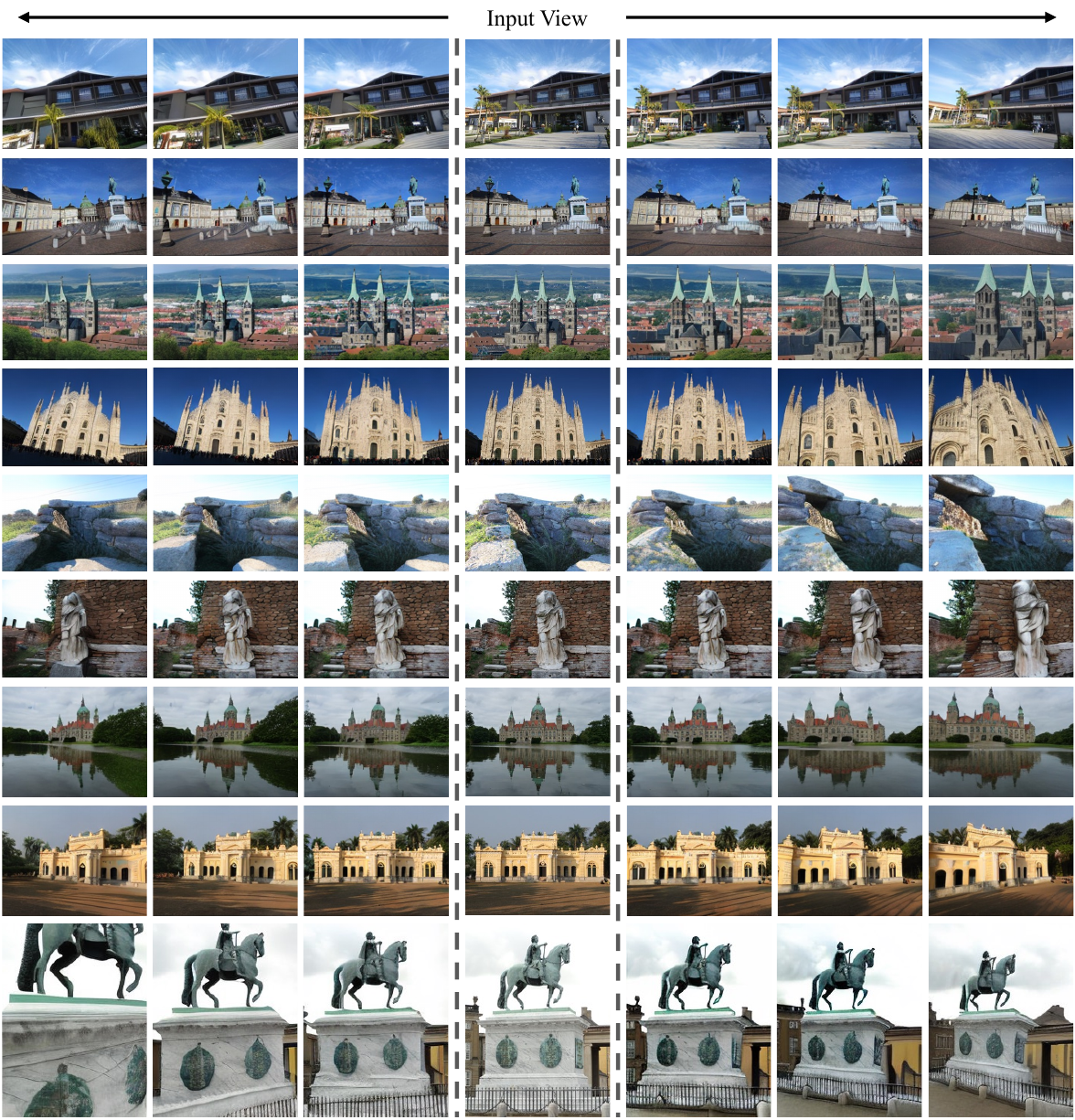} 
    \caption{\textbf{Qualitative results.} Additionally, we generate images using our method with samples from the MegaScenes \cite{megascenes} dataset, where the input view image in the middle column and continuous camera poses serve as inputs.}
    \label{fig:supple_5} 
\end{figure*}

In \namefig{}~\ref{fig:supple_2}, and \namefig{}~\ref{fig:supple_3}, we provide additional generation results for diffusion-based methods. 
\namefig{}~\ref{fig:supple_2} presents results generated using the MegaScenes \cite{megascenes} dataset, while \namefig{}~\ref{fig:supple_3} showcases results from the RE10K \cite{re10k} dataset.
As previously noted in prior research \cite{megascenes}, ZeroNVS \cite{ZeroNVS} fails to properly reflect viewpoint changes, often producing artifacts and inconsistencies across images. In contrast, our method generates more consistent images, where objects remain persistently visible across different viewpoints, and color variations are minimized, compared to MegaScenes and ZeroNVS.
Furthermore, to provide additional qualitative results, we present more examples from the RE10K and MegaScenes datasets in \namefig{}~\ref{fig:supple_4}, and  \namefig{}~\ref{fig:supple_5}. These results demonstrate the generalizability of our method across multiple datasets.
\clearpage
\clearpage
\clearpage
\clearpage
\section{Failure Cases}
\label{section_supp:eee}
\begin{figure}[h!]
    \centering
    \includegraphics[width=\linewidth]{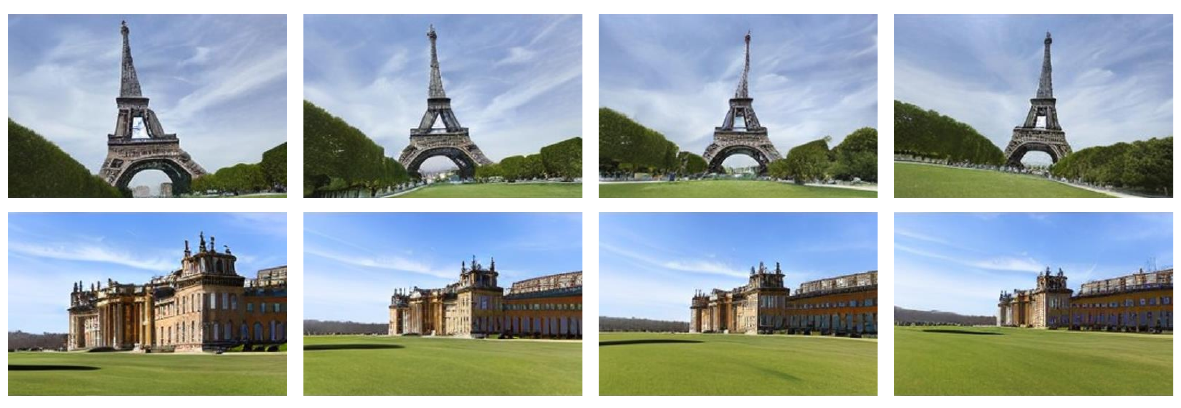} 
    \caption{\textbf{Failure cases.} \textit{Top}: an example that shows that the model struggles to accurately generate objects with thin structures, such as lines. \textit{Bottom}: an example that shows the model's tendency to repeatedly generate identical structural patterns as the viewpoint difference increases, reducing diversity.}
    \label{fig:supple_failure}
\end{figure}

Our method relies on novel view synthesis diffusion models that generate images from specific viewpoints. Consequently, any inherent limitations of these models are reflected in the generated results. Additionally, since our approach incorporates 3D warping, it becomes challenging to extract meaningful information as the viewpoint difference increases. 
In \namefig{}~\ref{fig:supple_failure}, we present examples of two key issues: (1) diffusion models exhibit distortions, particularly in thin structures such as lines, and (2) as the viewpoint difference increases, the generated results exhibit repetitive structures, leading to reduced diversity. This phenomenon can be attributed to the diminishing information provided by the warped images as the viewpoint increases. However, as discussed in the main paper, the problem introduced by the warping algorithm could be alleviated through an autoregressive approach, where a specific range is first generated and then iteratively used to synthesize subsequent ranges. This approach could serve as a future direction for overcoming the limitations of the warping algorithm.

\end{document}